\documentclass{article}

\usepackage{microtype}
\usepackage{graphicx}
\usepackage{subfigure}
\usepackage{booktabs} 

\usepackage{hyperref}
\usepackage{enumitem}

\usepackage[accepted]{icml2024}

\usepackage{amsmath,amsfonts,bm} \usepackage{amsthm}

\usepackage{thm-restate}
\newtheorem{theorem}{Theorem}

\def\rnn{{R}}

\usepackage{amssymb}

\newcommand\proofend{\hfill $\square$}

\def\eqref#1{equation~\ref{#1}}

\def\1{\bm{1}}

\def\F{{\mathcal{F}}}
\def\tF{{\tilde{\mathcal{F}}}}

\DeclareMathOperator{\vect}{vec}

\newcommand{\rvline}{\hspace*{-\arraycolsep}\vline\hspace*{-\arraycolsep}}

\DeclareMathAlphabet{\mathsfit}{\encodingdefault}{\sfdefault}{m}{sl}
\SetMathAlphabet{\mathsfit}{bold}{\encodingdefault}{\sfdefault}{bx}{n}

\newcommand{\R}{\mathbb{R}}
\newcommand{\C}{\mathbb{C}}

\newcommand{\diag}{\mathrm{diag}}

\usepackage{amsmath}
\usepackage{amssymb}
\usepackage{mathtools}
\usepackage{amsthm}

\usepackage[capitalize,noabbrev]{cleveref}

\usepackage[textsize=tiny]{todonotes}

\icmltitlerunning{Universality of Linear Recurrences Followed by Non-linear Projections}

\begin{document}

\twocolumn[
\icmltitle{Universality of Linear Recurrences Followed by Non-linear Projections: \\
{Finite-Width Guarantees and Benefits of Complex Eigenvalues}}



\icmlsetsymbol{equal}{*}

\begin{icmlauthorlist}
\icmlauthor{Antonio Orvieto}{xxx}
\icmlauthor{Soham De}{yyy}
\icmlauthor{Caglar Gulcehre}{zzz}
\icmlauthor{Razvan Pascanu}{yyy}
\icmlauthor{Samuel L. Smith}{yyy}
\end{icmlauthorlist}

\icmlaffiliation{xxx}{ELLIS Institute Tübingen, Max Planck Institute for Intelligent Systems, Tübingen AI Center, Tübingen, Germany}
\icmlaffiliation{yyy}{Google Deepmind}
\icmlaffiliation{zzz}{EPFL}

\icmlcorrespondingauthor{Antonio Orvieto}{antonio@tue.ellis.eu}

\icmlkeywords{Machine Learning, ICML}

\vskip 0.28in
]



\printAffiliationsAndNotice{}  

\begin{abstract}
Deep neural networks based on linear RNNs interleaved with position-wise MLPs are gaining traction as competitive approaches for sequence modeling. Examples of such architectures include state-space models~(SSMs) like S4, LRU, and Mamba: recently proposed models that achieve promising performance on text, genetics, and other data that require long-range reasoning. Despite experimental evidence highlighting these architectures' effectiveness and computational efficiency, their expressive power remains relatively unexplored, especially in connection to specific choices crucial in practice -- e.g., carefully designed initialization distribution and potential use of complex numbers. In this paper, we show that combining MLPs with both real or complex linear diagonal recurrences leads to arbitrarily precise approximation of regular causal sequence-to-sequence maps. At the heart of our proof, we rely on a separation of concerns: the linear RNN provides a lossless encoding of the input sequence, and the MLP performs non-linear processing on this encoding. While we show that real diagonal linear recurrences are enough to achieve universality in this architecture, we prove that employing complex eigenvalues near unit disk -- i.e., empirically the most successful strategy in S4 -- greatly helps the RNN in storing information. We connect this finding with the vanishing gradient issue and provide experiments supporting our claims.
\end{abstract}

\vspace{-6mm}

\textit{Note: The preliminary version of this manuscript~(ICML workshop version) only contains a subset of the results. Our follow-up~\cite{cirone2024theoretical} covers expressive power of gated SSMs such as Mamba~\cite{gu2023mamba}. }

\vspace{3mm}

\section{Introduction}

Attention \citep{vaswani2017attention} has supplanted LSTMs \citep{hochreiter1997long} and GRUs \citep{cho2014learning} as the dominant sequence-to-sequence mechanism for deep learning. However, a promising line of research sparked by the S4 model \citep{gu2021efficiently} has initiated a come-back of recurrent sequence models in simplified \textit{linear} form. A growing family of `state-space' models (SSMs) \citep{gu2021efficiently,hasani2022liquid,smith2023simplified, gu2022parameterization,li2022makes, orvieto2023resurrecting,gu2023mamba} has emerged which interleave recurrent linear complex-valued\footnote{The performance of non-selective SSMs trained \textit{from scratch } on classification tasks is greatly affected by the choice of number field: using complex diagonal recurrences greatly improves performance, see discussion in~\cite{gu2022parameterization,orvieto2023resurrecting} and, in particular, the S4D-Lin initialization. For Mamba on next-token prediction tasks, real numbers instead are effective. As shown by~\citet{amos2023never}, this is an effect due to a different learning paradigm which necessitates further investigations.} layers with other components such as position-wise multi-layer perceptrons (MLPs), gates~\citep{dauphin2017language}, residual connections \citep{he2016deep}, and normalization layers \citep{ioffe2015batch, ba2016layer}. 

Traditional SSMs achieve state-of-the-art results on the long-range arena~\citep{tay2020long} and show outstanding performance in various domain including vision~\citep{nguyen2022s4nd}, audio~\citep{goel2022sashimi}, biological signals~\citep{gu2021efficiently}, reinforcement learning~\citep{lu2023structured} and online learning~\citep{zucchet2023online}. SSMs, when augmented with a selectivity mechanism, are also successfully applied to text~\citep{katsch2023gateloop,gu2023mamba,de2024griffin}, and are sometimes used in combination with softmax attention~\citep{fu2023hungry, wang2023pretraining} or linear attention~\citep{peng2023rwkv, sun2023retentive, katsch2023gateloop, yang2023gated}. They gained significant interest in the literature due to two key factors. First, their computational complexity scales linearly in sequence length, while transformers scale quadratically \citep{vaswani2017attention, katharopoulos2020transformers}. Second, unlike LSTMs and GRUs, training can be efficiently parallelized~\cite{martin2017parallelizing}\footnote{At the current research state, effective parallelization in the time dimension is possible for recurrences which are linear in the state, and where the state-to-state transition is computationally light. Linear dense RNNs are parallelizable in principle but computationally heavy since unit operations involve dense matrix multiplications~\cite {smith2023simplified}.}.

\begin{figure*}
\vspace{-0mm}
    \centering \includegraphics[width=0.99\textwidth]{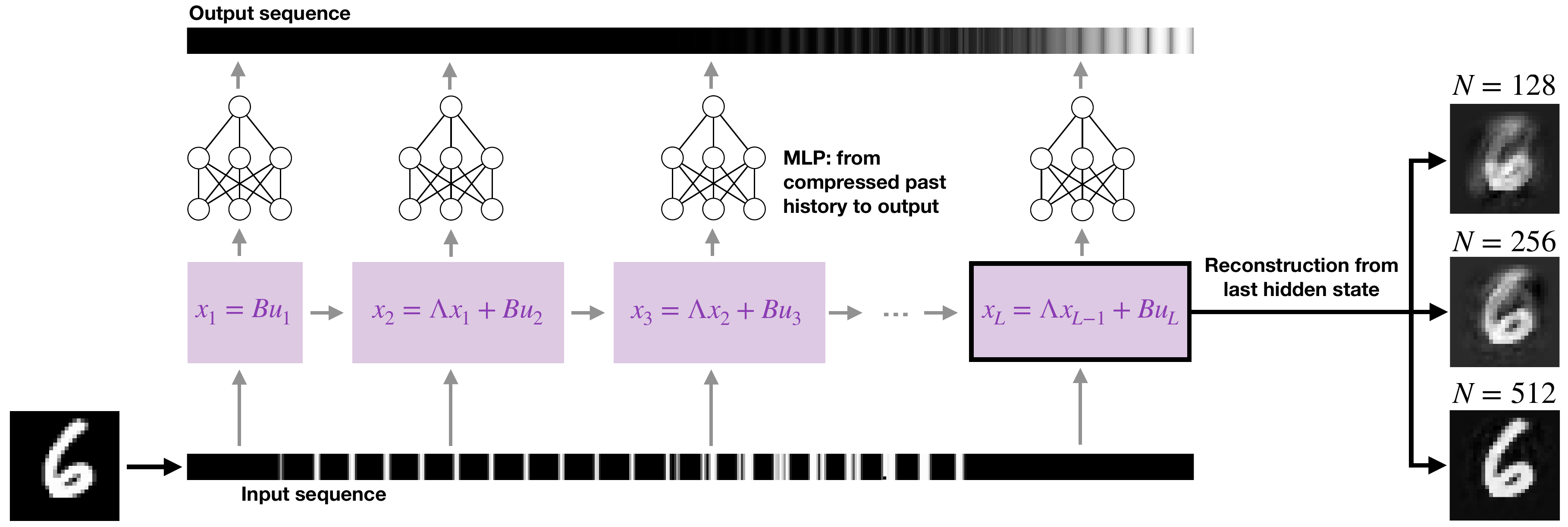}
    \vspace{-0mm}
\caption{Illustration of a Linear RNN + position-wise MLP on flattened MNIST~\citep{lecun1998mnist} digits. In our construction, the role of the linear RNN is to compress~(if possible) and store the input sequence into the hidden state: from hidden states one can recover past tokens using a linear transformation~(see \S\ref{sec:vandermonde}). As the hidden state size $N$ increases, the reconstructions becomes more and more faithful. The MLP~(same for all tokens) takes this representation as input, and is able to reproduce the action of any sufficiently regular sequence-to-sequence model~(see \S\ref{sec:MLP}). We provide additional insights and a thorough experimental evaluation and discussion in \S\ref{sec:discussion}.}
\label{fig:architecture}
\end{figure*}

The success of these models is surprising, since it was previously widely thought that non-linearities within the recurrence are necessary for RNNs to model complex interactions~\citep{schafer2006recurrent}. In this paper, we therefore step back and study the theoretical expressivity of models based on stacks of vanilla\footnote{For an analysis of selective SSMs such as Mamba~\cite{gu2023mamba} and Griffin~\cite{de2024griffin}, please check our follow-up~\cite{cirone2024theoretical}.} linear complex-valued RNNs interleaved with position-wise MLPs. In contrast to non-linear RNNs for which we have several expressivity results, from Turing-completeness~\citep{Chung21,Siegelmann92} to universality~\citep{schafer2006recurrent, hanson2020universal}, less is known for architectures based on linear recurrences, with nonlinearities placed outside of the sequential processing.

\vspace{-3mm}
\paragraph{Results on SSMs expressivity.} The Mütz-Szász Theorem~\citep{muntz1914approximationssatz, szasz1916approximation} can be used to show that linear diagonal real-valued RNNs can approximate causal convolutions of the input with an arbitrary filter in the width limit~\citep{li2022approximation}. However, linear filtering alone cannot provide satisfactory approximations of non-linear causal sequence-to-sequence maps. Yet, interestingly, \citet{hanson2019universal} showed that staking an exponential~(in the sequence length) number of filters, chained with ReLUs, can approximate arbitrary time-homogeneous nonlinear causal operators. Recently, in the more realistic finite-depth regime of SSMs, \citet{wang2023state} proved that interleaving five linear RNNs with nonlinearities leads, in the width limit, to the approximation of any continuous mappings from an input sequence to a scalar target. This result uses the Kolmogorov-Arnold Theorem~\citep{kolmogorov1957representation}. Approximating nonlinear causal sequence-to-sequence mappings~(in the sense of Definition~\ref{def:seq}) is, of course, harder: \citet{wang2023state} provided an interesting connection with Volterra-series expansions~\citep{boyd1985fading} of time-invariant causal operators: multiplying the output of $K$ infinitely wide linear RNNs provides a $K$-th order approximation of smooth non-linear sequence to sequence mappings. While this result establishes a connection to the rich literature in dynamical systems, it concerns a specific architecture design strategy~(product of $K$ RNN outputs), which is far from SSMs practice. Further, the discussion in~\citet{wang2023state} does not characterize the effect of choosing complex-valued recurrences~(as crucially done in most SSMs, including S4)  on expressivity. In this paper, we adopt a different strategy compared to~\citet{wang2023state}, and aim instead at characterizing the information content in the RNN hidden state, further studying how this information is processed by the MLP across timestamps.

\vspace{-3mm}
\paragraph{Contributions.} We prove that a single linear diagonal RNN layer followed by a position-wise MLPs can approximate arbitrarily well any \emph{sufficiently regular}~(see Definition~\ref{def:seq}) nonlinear sequence to sequence map over finite length sequences. Our key insight is that, if the linear RNN can preserve a lossless compressed representation of the entire sequence seen so far, the MLP has access to all of the information in the input. Since MLPs are universal approximators~\cite {pinkus1999approximation}, they can process this information to achieve any desired mapping~(see Fig.~\ref{fig:architecture}). Along our reasoning, we give several insights on the effects of initialization and characterize the role of complex numbers on memory.
\vspace{-3mm}
\begin{enumerate}[left=0.1em, itemsep=0.5pt]
    \item In \S\ref{sec:vandermonde}, using the properties of Vandermonde matrices, we show that at a fixed timestamp $k$ one can precisely (zero error) reconstruct the value of each seen input token from the hidden state of a random diagonal linear RNN. We discuss how wide the RNN should be for this property to hold, and how the result adapts to the setting of compressible inputs. These results follow from straightforward linear algebra arguments but provide great insights into RNN memorization mechanisms. Our discussion has deep links to the HiPPO theory~\citep{gu2020hippo} but does not require structured matrices at the cost of stronger assumptions on the RNN width~(\S\ref{sec:hippo}). 
    \item In \S\ref{sec:MLP}, we prove our claim: linear RNNs followed by position-independent MLPs with one hidden layer can approximate regular sequence-to-sequence maps. Starting from the results of \S\ref{sec:vandermonde}, we use Barron's theory~\citep{barron1993universal} to provide guarantees on the MLP width. This result involves technical steps such as the computation of the Barron constant of the interpolation of non-linear maps. Crucially, we find that the MLP width is affected by the ease of reconstruction from the linear RNN state, quantified by the condition number of the reconstruction map, which is heavily dependent on the RNN initialization.
    \item In \S\ref{sec:discussion}, we leverage our framework to understand some interesting features of basic SSMs such as S4~\citep{gu2021efficiently} and the LRU~\citep{orvieto2023resurrecting} -- such as the need, e.g. when training from scratch on the Long-range Arena~(LRA)~\citep{tay2020long}, for complex-valued recurrences and initialization near the unit circle in $\C$. In the same section,  we validate our claims and intuition on initialization on synthetic datasets and on LRA tasks~\citep{tay2020long}.
\end{enumerate}
\vspace{-2mm}
Due to space limitations, proofs and additional related works are presented in the appendix.

\section{Preliminaries}
\label{sec:pre}
Consider length-$L$ sequences of real $M$-dimensional inputs: $\bar v=(v_i)_{i=1}^L\in\mathcal{V}\subseteq\R^{M\times L}$. We often refer to each $v_i$ as a ``token''\footnote{In formal NLP~\cite{cotterell2023formal}, $\mathcal{V}$ is a finite vocabulary. Here, we discuss the setting where $\mathcal{V}$ is possibly dense in $\R^{M\times L}$. }. A \textit{sequence-to-sequence map} is a deterministic transformation of input sequences that produces output sequences of the same length: $\bar y=(y_i)_{i=1}^L\in\mathcal{Y}\subseteq\R^{S\times L}$. We say that a sequence-to-sequence map is \textit{causal} if for every $k$, $y_k$ is blind to the tokens $(v_i)_{i=k+1}^L$.  

\begin{restatable}[Sequence-to-sequence]{definition}{seq-to-seq}
\label{def:seq}
A causal sequence-to-sequence map with length-$L$ sequential $M$-dimensional inputs $\bar v=(v_i)_{i=1}^L\in\mathcal{V}\subseteq\R^{M\times L}$ and length-$L$ sequential $S$-dimensional outputs $\bar y=(y_i)_{i=1}^L\in\mathcal{Y}\subseteq\R^{S\times L}$ is a sequence of non-linear continuous functions $T=(T_k)_{k=1}^L$, $T_k:\R^{M\times k}\to\R^S$ $\forall k\in[L]$. $T$ acts as follows: 
\vspace{-2mm}
$$(v_i)_{i=1}^L\overset{T}{\mapsto} (y_i)_{i=1}^L \ \ , \ \ \text{s.t.} \ \ y_k = T_k((v_i)_{i=1}^k).$$
\end{restatable}
\vspace{-2mm}
We are going to assume without restating this that each $T_k$ is a Barron function~(Def.~\ref{def:barron}), with a well-defined integrable Fourier transform~(used in Thm.~\ref{thm:comb_barron}). We consider approximations $\hat T$ to $T$ using a neural network. Schematically:
$$\hat T = g \circ \rnn \circ e,$$
\vspace{-5mm}
where:
\begin{itemize}
\item $e:\R^M\to\R^H$ is a linear embedding layer with biases, acting tokenwise. We denote the encoded tokens sequence $\bar u = (u_i)_{i=1}^L\in\R^{H\times L}$, where $u_k = e(v_k)\in\R^H$, for all $k\in[L]$.
    \item $\rnn$ is a linear RNN processing the encoded input tokens $(u_i)_{i=1}^L$ producing a sequence of $N$-dimensional hidden states $(x_i)_{i=1}^L\in\R^{N\times L}$.
    \item $g:\R^{N}\to\R^S$ is a non-linear function, acting tokenwise, parametrized by an MLP. We have that $\hat{y_k} = g(x_k)\in\R^S$, for all $k\in[L].$
\end{itemize}
\vspace{-3mm}
The combination of $g$ with $\rnn$ and $e$, which we denote as $g \circ  \rnn \circ e$, produces outputs $\hat{\bar{y}} = (\hat y_i)_{i=1}^L\in\R^{P\times L}$ which we desire to be close to $\bar y = T(\bar v)$.

The Linear RNN $R$ processes inputs recurrently. This operation is parametrized by matrices $A$ and $B$. 
\begin{restatable}[Linear RNN]{definition}{lrnn}
$\rnn_{A,B}:\R^{H\times L}\to\R^{N\times L}$ processes an (encoded) sequence $\bar u = (u_i)_{i=1}^L$ producing an output sequence of hidden states $\bar x=(x_i)_{i=1}^L\in\R^{N\times L}$ by means of the following recursion:
\begin{equation}
\label{eq:lrnn}
   x_{k} = A x_{k-1} + B u_k, 
\end{equation}
where $A\in\R^{N\times N}$, $B\in\R^{N\times H}$, and $x_{0}=0\in\R^N$.
\end{restatable}
\vspace{-2mm}
\paragraph{Diagonal Linear RNNs.} Linear RNNs have been shown to be very successful token-mixing components in deep architectures such as SSMs~\citep{gu2021efficiently, orvieto2023resurrecting}. A crucial feature of SSMs -- making them appealing compared to non-linear variants such as LSTMs~\citep{hochreiter1997long} or GRUs~\citep{cho2014learning}-- is that the forward pass is cheap to compute by means of parallel scans~\citep{blelloch1990prefix}. At the root of such fast computation is the diagonalizability property of linear RNNs.\footnote{In principle, also non-diagonal linear RNNs can be parallelized. Yet, parallelizing non-diagonal RNNs involves multiplying dense matrices. To achieve a speedup over recurrent computation on modern hardware, the linear RNN has to be diagonal.} Indeed, over the space of $N\times N$ non-diagonal real matrices, the set of non-diagonalizable~(in the complex domain) matrices has measure zero~\citep{bhatia2013matrix}. Hence, up to arbitrarily small perturbations, $A$ is diagonalizable over the complex numbers, i.e. one can write $A = Q\Lambda Q^{-1}$, where $\Lambda = \diag(\lambda_1, \dots, \lambda_{N}) \in \mathbb{C}^{N \times N}$ gathers the eigenvalues of $A$, and the columns of $Q\in\C^{N\times N}$ are the corresponding eigenvectors. 
\begin{align*}
    & x_k= Q\Lambda Q^{-1} x_{k-1} + B u_k\\
    \implies & (Q^{-1}x_{k}) = \Lambda (Q^{-1} x_{k-1}) + (Q^{-1} B) u_k
\end{align*}
By renaming $x_k\leftarrow Q^{-1}x_{k}$ and $B\leftarrow Q^{-1}B$, one arrives at the complex-valued diagonal recursion 
\begin{equation}
\label{eq:ldrnn}
   x_{k} = \Lambda x_{k-1} + B u_k.
\end{equation}
The corresponding map $\rnn_{\Lambda,B}:\R^{N\times L}\to\C^{N\times L}$ is such that $\rnn_{A,B} = Q\circ \rnn_{\Lambda,B}$, where $Q$ is applied tokenwise. Since we are interested in architectures of the form $\hat f = g \circ \rnn \circ e$, the linear transformation $Q$ can be merged into $g$, at the price of having inputs for $g$ with \textit{doubled dimension}~(real and imaginary parts\footnote{In practice, some SSM variants drop the imaginary part.}): $g:\R^{2N}\to\R^P$.
Without loss in generality~(see also \citet{li2022approximation}), we therefore assume from now on our linear RNNs are in diagonal form. Morover, since the linear transformation $Q$ can be merged with the output projection, we consider -- as done in S4 and LRU -- initialization directly on the eigenvalues, and \textit{not passing through diagonalization} of a Glorot-initialized dense matrix $A$. Note that while the role of complex numbers in this setting is clearly motivated by the construction, we later show in \S\ref{sec:discussion} that using complex eigenvalues induces peculiar memorization properties.
\vspace{-1mm}

\paragraph{MLPs and universality.} MLPs with one hidden-layer~(1HL-MLPs) are universal non-linear function approximators~\citep{barron1993universal, pinkus1999approximation}. In this paper, we consider $g$ parametrized by a 1HL-MLP. $g$ is a proxy for a general non-linear function $f$, which is regular in the sense that it is \textit{not oscillating too quickly} -- meaning that its Fourier transform $F(\omega)$ exhibits fast decay as $\|\omega\|\to\infty$.

\begin{restatable}[Barron function]{definition}{barron}
\label{def:barron}
Let $\mathcal{F}(\omega)$ be the Fourier transform of $f:\R^n\to\R$. $f$ belongs to the Barron class if $C_f = \int_{\R^n} \|\omega\|_2 |\mathcal{F} (\omega)|d\omega<\infty$.
\end{restatable}

\vspace{-2mm}
We have the important result due to~\cite{barron1993universal}.

\begin{restatable}[Universality of 1HL-MLPs]{theorem}{barronthm}
\label{thm:barron_thm}
Consider $g(x)$ parametrized by a 1HL-MLP (with $D$ hidden neurons): $g(x) = \sum_{k=1}^D \tilde c_k\sigma(\langle \tilde a_k , x\rangle + \tilde b_k)+ \tilde c_0$, where $\sigma$ is any sigmoidal function\footnote{$\lim_{x\to-\infty} \sigma(x) = 0$ and $\lim_{x\to\infty} \sigma(x) = 1$.}. Let $f:\R^n\to\R$ be continuous with Barron constant $C_f$. If $D\ge 2 r^2 C_{f}^2 \epsilon^{-2}$, then there exist parameters such that $\sup_{\|x\|\le r} |f(x)-g(x)|\le\epsilon$.
\end{restatable}
\vspace{-2mm}
Note that ReLU activations also work in the setting of the theorem above, up to a factor $2$ in the bound, since $\text{ReLU}(x)-\text{ReLU}(x-1)$ is sigmoidal.
\begin{restatable}[Multidimensional Output]{remark}{barron_rmk_multi}
\label{rmk:multidim_barron}
The result above is stated for the scalar output case. The $S$-dimensional outputs result can be easily derived by stacking neurons in the hidden layer, that becomes of dimension $D\leftarrow DS$. Therefore if $D\ge 2 r^2 C_{f}^2 S\epsilon^{-2}$, then $\sup_{\|x\|\le r} \|f(x)-g(x)\|_1\le\epsilon S$ (since errors accumulate). Let us then call $\epsilon \leftarrow \epsilon S$ the desired accuracy; the number of neurons needed to achieve that is $D\ge 2 r^2 C_{f}^2 S^3\epsilon^{-2}$.
\end{restatable}

\section{Universality Result}
\label{sec:universal}
In this section, we show that, as the model $\hat T = g \circ R \circ e$ grows in width, there exist network parameters such that $\hat T \approx T$. We state this result informally below.

\begin{restatable}[Universality]{theorem}{main_thm}
\label{thm:universal}
Let the inputs set $\mathcal{V}\subset\R^{M\times L}$ be bounded and $\epsilon>0$ be the desired accuracy in approximating $T$. Let $\rnn$ be a diagonal linear real or complex RNN with width $N\ge \dim(\mathcal V)$ and let the MLP width $D\ge O(L/\epsilon^2)$. Then, $\hat T = g \circ R \circ e$ approximates pointwise $T$ with error $\epsilon$:  $$\sup_{\bar v\in \mathcal{V}}\|\hat T (\bar v)-T(\bar v)\|\le \epsilon.$$
\end{restatable}
\vspace{-2mm}
Here, by $\dim(\mathcal{V})$ we mean the \textit{vector-space dimension} of $\mathcal{V}$. In the worst-case scenario where inputs are not structured, $\dim(\mathcal V) = LM$. In practice, we observe that one can work with smaller dimensions in the hidden state~(Fig.~\ref{fig:MNIST_PF_rec_MLP}).
The proof comprises two steps:
\vspace{-3mm}
\begin{itemize}[left=0.1em, itemsep=1pt]
    \item Step 1: Linear RNNs can perform lossless compression~(\S\ref{sec:vandermonde}): from the RNN output $x_k$ one can perfectly~(no error) reconstruct the input $(v_i)_{i=1}^k$, assuming $N$ is large enough~($N\ge \dim(\mathcal V)$).
    \item Step 2: The MLP head $g$ on top of $x_k$ can reconstruct the ground-truth map $T_k$: $T_k((v_i)_{i=1}^k)\simeq g(x_k)$, assuming the number of hidden MLP neurons $D$ is large enough~($D\ge O(L/\epsilon^2)$).
\end{itemize}
\vspace{-2mm}
While step 2 is mainly technical, step 1 -- dealing specifically with the RNN -- is less involved and leads to valuable insights into the architecture.
\begin{restatable}{remark}{compression}
\label{rmk:compression}
A few comments on the result are needed:
\vspace{-3mm}
\begin{itemize}[left=0.1em, itemsep=1pt]
    \item In Thm.~\ref{thm:universal}, both the size of the RNN and the MLP agree with basic information theory reasoning: (a) if inputs are random and unstructured, the hidden state cannot perform compression: the RNN has to store $\mathcal{O}(L)$ reals, requiring $\mathcal{O}(L)$ hidden dimensions; (b) the MLP size is $\mathcal{O}(L)$ since, in the worst-case, it has to model $L$ distinct maps $T_1, T_2,\dots, T_L$~(Def.~\ref{def:seq}). This complexity can be reduced by assuming temporal smoothness~(cf. \S\ref{sec:hippo}).
    \item The setting of tokens living in a finite set~\cite{cotterell2023formal} is drastically different and not discussed in this work. An interesting result using a construction derived from the first version of this paper can be found in~\citet{ding2023recurrent}. Additional recent results can be found in~\citet{jelassi2024repeat}.
    \item A simple application of Barron's Theorem~(Thm.~\ref{thm:barron_thm}) does not suffice to prove Step 2. In SSMs, the same MLP function is applied at each timestamp, ant therefore has to model the interpolation of $T_1, T_2,\dots, T_L$. Adapting Barron's theory to this time-dependent setting is our main technical contribution.
\end{itemize}
\end{restatable}
This section is dedicated to the proof of Thm~\ref{thm:universal}, with the main intuition relevant for our discussion is presented in \S\ref{sec:main_idea}. Valuable insights on initialization and role of complex numbers can be derived from our proof strategy. These are discussed in \S\ref{sec:discussion}. 
\subsection{Linear RNNs can perfectly memorize inputs}
\label{sec:vandermonde}
We first state the main result of this subsection.
\begin{restatable}[Power of Linear RNNs, informal]{theorem}{pow_rnns}
\label{thm:perfect}
Under no assumption on the encoded inputs set $\mathcal{U}\subseteq \R^{H\times L}$, if the hidden dimension scales with $HL$, then linear RNN-based processing comes with \textbf{no information loss}: inputs $(u_i)_{i=1}^k$ can be perfectly recovered from $x_k$, for any $k=1,2,\dots, L$. If $\dim(\mathcal{U})=:P<HL$ -- i.e. if $\mathcal{U}$ is linearly \textbf{compressible} -- the hidden dimension can be reduced to $\mathcal{O}(P)$.
\end{restatable}
\vspace{-2mm}
We encourage the reader to go through this subsection to get a proof idea. Insights and practical considerations will be then addressed thoroughly in~\S\ref{sec:discussion}. In \S\ref{sec:hippo}, we compare this simple result with HiPPO theory~\citep{gu2020hippo}: a framework that provides precise guarantees on the approximation error of \textit{structured} linear RNNs at any width.

In this subsection, with ``input'' we refer to the encoded sequence $\bar u = e(\bar v)\in\R^{H\times L}$. Note that, as typically $H\ge M$, accurate reconstruction of $\bar u$ implies accurate reconstruction of $\bar v$. We start by unrolling Eq.~(\ref{eq:ldrnn}): $x_{1} = Bu_1$,  $x_2 = \Lambda Bu_1 + Bu_2$, $x_3 = \Lambda^2Bu_1 + \Lambda Bu_2 + B u_3$, and 
\begin{align}
    \label{eq:lin_rnn_unroll}
    x_k =\sum_{j=0}^{k-1} \Lambda^jBu_{k-j}.
\end{align}
It is easy to realize that this operation stacks convolutions of the sequence $ B \bar u\in\C^{N\times L}$ with $N$ independent one-dimensional filters parametrized by $\Lambda=\diag(\lambda_1,\dots,\lambda_N)$, and with form $(1, \lambda_i,\lambda_i^2,\dots, \lambda_i^{L-1})$ for $i=1,2,\dots N$. 
\subsubsection{Main idea}
\label{sec:main_idea}
Let $H=M=1$, the encoder $e$ be the identity, and $B = (1,1,\dots, 1)^\top$. Then, Eq.~(\ref{eq:lin_rnn_unroll}) can be written as
\begin{equation}
    x_k   =
    \begin{pmatrix}
    \lambda_1^{k-1}&\lambda_1^{k-2} &\cdots& \lambda_1&1\\
    \lambda_2^{k-1}&\lambda_2^{k-2} &\cdots& \lambda_2&1\\
    \vdots&\vdots &\ddots&\vdots&\vdots\\
    \lambda_N^{k-1}&\lambda_N^{k-2} &\cdots& \lambda_N&1\\
    \end{pmatrix}
    \begin{pmatrix}
    u_1 \\ u_{2} \\ \vdots \\ u_k
    \end{pmatrix}= V_k u_{1:k}^\top.
    \label{eq:vandermonde}
\end{equation}
where $u_{1:k}= v_{1:k} = (v_i)_{i=1}^k \in\R^{1\times k}$, and $V_k$ is a Vandermonde matrix. As long as $N\ge k$, we can hope to recover $u_{1:k}$ by pseudoinversion of the Vandermonde:
\begin{equation}
    v_{1:k}^\top = u_{1:k}^\top = V_k^+ x_k.
\end{equation}
Indeed, note that if $N=k$~(number of equations $=$ number of unknowns), the matrix $V_k$ is invertible under the assumption that all $\lambda_i$ are distinct, since~(see e.g.~\cite{bhatia2013matrix}): 
$\det(V_k) = \prod_{1\le i< j\le k}(\lambda_i-\lambda_j)\ne 0$. If $N>k$ then under again the assumption that eigenvalues are distinct, the linear operation on the hidden state $V_k^+ x_k$ produces the input sequence without errors. To make this procedure work in the simplest setting, a sufficient condition is $N\ge L$. We summarize below.

\begin{restatable}[Bijectivity]{proposition}{main_simple}
Let $M=1$, $H=1$ and let the encoder $e$ be the identity. Consider input projection $B = (1,1,\dots, 1)^\top$ and recurrent matrix $\Lambda=\diag(\lambda_1,\dots,\lambda_N)$ with $\lambda_i\ne\lambda_j$ for all $i\ne j$. Fix $k\in[L]$, let $R_{\Lambda,B}^k:\R^{H\times k}\to\R^N$ denote the map $(u_{i})_{i=1}^k\mapsto x_k$. If $N\ge L$, $R_{\Lambda,B}^k$ is bijective\footnote{Note that this maps the input sequence to the last hidden state.}, with linear inverse.
\end{restatable}
\vspace{-3mm}
The reconstruction map~(parametrized by $V_k^+$) is time-variant and has \textit{time-dependent} output dimension $\R^k$. Since the RNN output we consider in this paper is a \textit{time-independent MLP} head $g$, it is essential for the hidden state to also contain information about \textbf{time}. This is trivial to achieve with linear RNNs: consider $e:\R^M\to\R^{M+1}$ such that $v_k$ is mapped to $u_k = (1, v_k)$ for all $k$. Let $B = ((1,0,0,\dots, 0),(0,1,1,\dots,1))^\top$. Consider $\lambda_0=1$. Then, the first hidden state dimension will coincide with the index $k$, since $x_{k,0} = x_{k-1,0} + 1$ for all $k$. The construction above, allows the RNN to count the step it is on. In \S\ref{sec:MLP}, we will use this fact to conclude the theoretical discussion. 

\vspace{-3mm}
\paragraph{Multidimensional setting.} Since the RNN is linear, one can design $M$ independent RNNs acting on separate input dimensions. Such RNNs can be combined into one single block using a properly designed $B$ matrix.


\subsubsection{RNNs can compress inputs, when possible}
\label{sec:sparse}
What we discussed in Theorem~\ref{thm:perfect} is a \textbf{worst-case setting} that requires the RNN hidden state $N$ to be of size proportional to the input sequence $L$. This is not surprising since we made no assumptions about the inputs $\bar u$: it is indeed not possible in general to store $\mathcal{O}(L)$ real numbers in a vector of size smaller than $\mathcal{O}(L)$ by means of a linear Vandermonde projection -- unless such inputs are structured~(cf. \S\ref{sec:hippo}). However, the RNN hidden state dimension scales efficiently under low-dimensionality, which we refer to as \textit{sparseness} in relation to sparse coding~\citep{lewicki2000learning}.

\begin{restatable}[Low-dimensional, a.k.a \textit{sparse}, inputs]{assumption}{sparse} Let $\Psi = (\psi^i)_{i=1}^P$, with $\psi^i\in\R^{M\times L}$ for all $i=1,2,\cdots, P$ be an ordered list of basis functions. For every input sequence $\bar v$, there exist coefficients $\alpha^v:=(\alpha^{v}_i)_{i=1}^P\in\R^P$ such that $\bar v  = \sum_{i=1}^P \alpha^{v}_i\psi^i$. We do not assume such decomposition is unique, nor to have access to the basis functions.
\label{ass:sparse}
\end{restatable}
\vspace{-2mm}
Let us discuss this assumption in the one-dimensional setting $M=1$. Definitions and results carry out effortlessly to the multidimensional setting. In matrix form, Assumption~\ref{ass:sparse} can be written as: there exists a real matrix $\Psi\in\R^{L\times P}$ such that, for all $\bar v\in\mathcal{V}\subset\R^{1\times L}$ there exists $\alpha^v\in\R^P$ such that $\bar v^\top = \Psi \alpha^v$. Let $\Psi_k\in\R^{k\times P}$ denote the first $k$ columns of $\Psi$. The last equation implies $v_{1:k}^\top = \Psi_k \alpha^v_k$, where $\alpha_k$ is one among the estimates of the coefficients $\alpha^v$ holding until iteration $k$~(e.g. one frequency component might be visible only after a specific $k$). Eq.~(\ref{eq:vandermonde}) then implies:
\begin{equation}
    x_k  = V_k v_{1:k}^\top = V_k\Psi_k \alpha^v_k,
\end{equation}
Under the assumption that $\Gamma_k := V_k\Psi_k$ is full rank, $\alpha^v_k = (\Gamma_k^\top \Gamma_k)^{-1} \Gamma_k^\top x_k$. We therefore get:
\begin{equation}
    v_{1:k} = \Omega_k x_k, \quad \Omega_k:=\Psi_k(\Gamma_k^\top \Gamma_k)^{-1} \Gamma_k^\top \in\C^{k\times N}.
\end{equation}

\vspace{-4mm}
\paragraph{General definition of $\boldsymbol{\Omega_k}$.} In the non-sparse setting, it is easy to see that $\Omega_k=V_k^+$. Therefore, from this point in the paper \textit{we will keep denoting as $\Omega_k$ the linear reconstruction map}. To make $\Omega_k$ act on real numbers, we consider instead its real/imaginary inputs representation in $\R^{2N\times N}$.

\vspace{-4mm}
\paragraph{$\boldsymbol{\Psi}$ may be unknown.} One does not need to assume a specific $\Psi$ a priori: under input sparsity, we can guarantee approximation with a learned linear map on the state, which may be task-dependent. Even if the input is strictly-speaking not sparse, not all information might be relevant for prediction. In a way, sparsity mathematically quantifies the belief that the information content relevant to a certain task has a lower dimension compared to the original signal.

\subsubsection{Comparison with HiPPO}
\label{sec:hippo}

\begin{figure*}[h]
    \centering
    \includegraphics[width=0.86\textwidth]{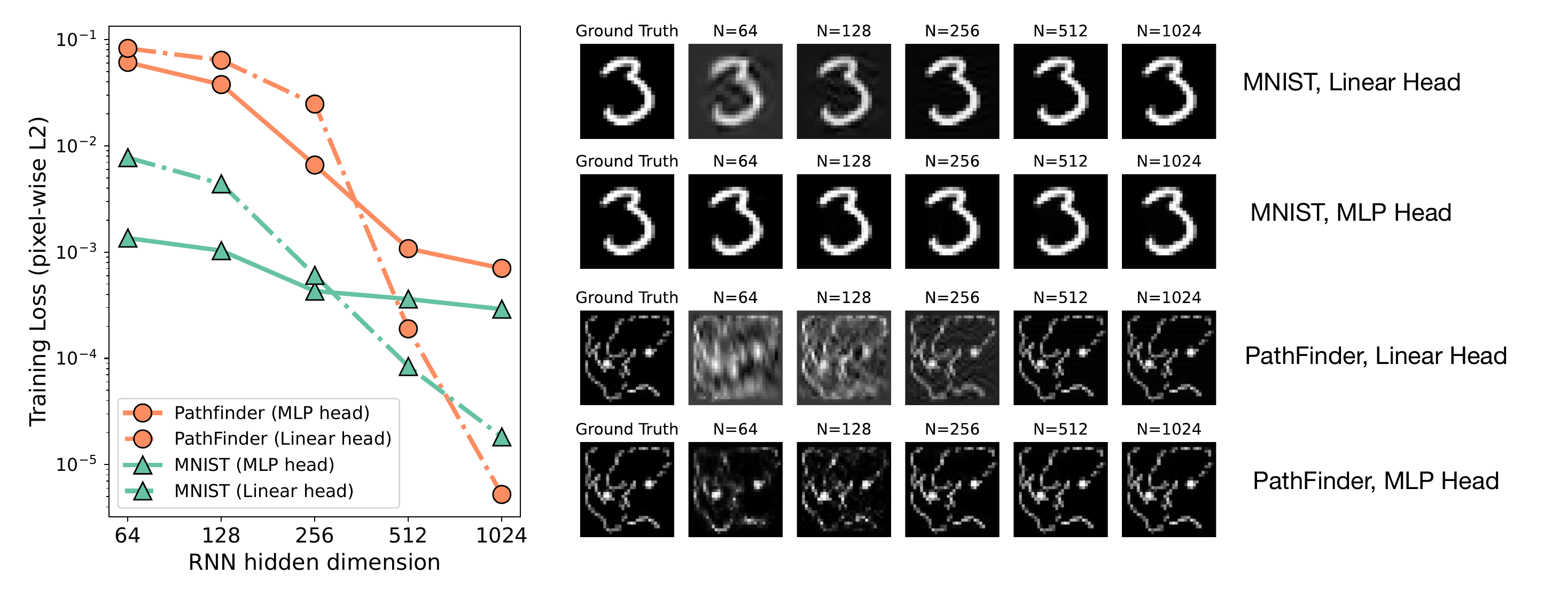}
    \vspace{-2mm}
    \caption{Reconstruction of MNIST digits and PathFinder data~\citep{tay2020long} Using a trained RNN + MLP or linear reconstruction decoder \textit{from the last hidden state} $x_L$. Plotted is average L2 pixel-wise norm~(mean of 3 runs). All parameters are trained, hyperparameters are tuned. $r_{\min} = 0.9$, $r_{\max}=0.999$ are found to be best for initialization of the RNN~(in line with~\cite{gu2021efficiently}). For large hidden dimension, linear reconstruction is successful. For smaller hidden dimension, non-linear reconstruction becomes necessary.}
    \label{fig:MNIST_PF_rec_MLP}
    \vspace{-2mm}
\end{figure*}

HiPPO theory~\citep{gu2020hippo} describes the compression properties of continuous-time linear RNNs of the form $\dot x_t = A x_t + Bu_t$, where $u:\R\to\R$ is a \textit{smooth input} and $A\in\R^{N\times N}, \ B\in\R^N$ have a specific fixed form~(i.e. are \textit{structured}). A common choice for $A$ and $B$ implements the LegS operator. At time $t$, the LegS operator maps the input to the $N$ polynomial coefficients $c_i(t)$ such that $g^{(t)} = \sum_{i=1}^N c_i(t) p_i^{(t)}$ is the best possible polynomial approximation of $u_{[0,t]}$ under the uniform measure in time on the interval $[0,t]$. If the inputs are Lipschitz continuous~(smooth), they can indeed be approximated by an $N$-degree polynomial: the error is then simply the sum of the small polynomial coefficients above degree $N$: $|u_{[0,t]}-g^{(t)}|^2 = O( t^2 \ell^2 / N)$. This bound is due to the fundamental properties of polynomials and can be considered independently from the HiPPO framework: being $\ell$-Lipschitz implies an error $O(t^2 \ell^2 / N)$ from inputs being a projection on a specific basis on $N$ orthogonal polynomials. In particular, note that assuming the error is $O( t^2 \ell^2 / N)$ from being described by the first $N$ coefficients is actually a weaker requirement compared to $\ell$-Lipshitzness. This is linked with sparseness : $u_{[0,t]} = \sum_{i=1}^N c_i(t) p_i^{(t)}$ can be framed as Assumption A.
While HiPPO theory, in contrast to this paper, allows for precise error estimates, our approach is more general. We show that it is not needed to fix an input measure or to hand-pick $A$, $B$: random initialization on a ring can already lead to successful reconstruction in practice~(Fig.~\ref{fig:MNIST_PF_rec_MLP}). Our worst-case result holds if $N\ge L$, \textit{i.e. for the general non-smooth input setting}. Furthermore, as showcased above, Assumption A can be linked to polynomial regression, and therefore describes a general property -- at the price of missing error quantifications. This can be resolved: for a fixed basis $\Psi$, it is possible to provide bounds using the famous lemma of~\citet{johnson1986extensions}: a similar analysis is used to quantify the error of randomized signatures in Rough Paths Theory~\citep{cuchiero2021expressive, compagnoni2023effectiveness}.\\
Compared to HiPPO, an advantage of our simplified~(yet less precise) approach is that we can easily move to the question of expressivity, which we explore next. Further, our non-structured approach leads to additional insights into the role of complex eigenvalues, which we explore in \S\ref{sec:discussion}.

\subsection{MLP Reconstruction}
\label{sec:MLP}
Recall that our goal is to approximate the object $T=(T_i)_{i=1}^L$, that maps $(v_i)_{i=1}^L\overset{T}{\mapsto} (y_i)_{i=1}^L$ such that $y_k = T_k((v_i)_{i=1}^k)$ for all $k=1,2,\dots, L$. In the usual SSMs pipeline, this approximation takes the form $\hat T = g \circ \rnn \circ e$, where compositions are tokenwise~(MLP $g$ and embedding $e$ applied to each token), and $R$ is a linear diagonal RNN, mixing tokens in the temporal direction.\\
Let $x_k$ be the $k$-th RNN output. In the settings described in the last subsection, for every $k\in\{1,2,\dots, L\}$ there exists a linear function $\Omega_k'$~(identity on the first coordinate $x_{k,0}=k$, and $\Omega_k$ on the other coordinates) such that $\Omega_k' x_k = (k,(v_i)_{i=1}^k)$. Note that under no assumption on the set of inputs, $\Omega_k$ coincides with the Vandermonde inverse $V_k^+$. The last operation, $(k,v_{1:k})\mapsto y_k$ needs to be performed by the \textit{time-independent} map $g$. 

\vspace{-3mm}
\paragraph{Step 1: fixed timestamp.} Let us focus on getting the MLP to approximate $T_k$, $k$ fixed, from the RNN output.  $g$ has to satisfy $g(x_k) \overset{!}{=} T_k(\Omega_k(x_k)) =: f_k(x_k)$.
The following result~(based on Rmk.\ref{rmk:multidim_barron}) bounds the width of the MLP $g$ for approximation of this quantity.
\begin{restatable}[MLP, single timestamp]{proposition}{mainMLPsing}
\label{prop:single_timestamp}
The number of hidden neurons in the MLP $g$ sufficient to approximate $f_k:=T_k\circ\Omega_k$ at level $\epsilon$ from the RNN hidden state $x_k$ is bounded by $4 r^2 C_{T_k}^2 \|\Omega_k\|^2_2 S^3\epsilon^{-2}$ where $r$ bounds the hidden state magnitude, and $C_{T_k}$ is the Barron constant of $T_k$.
\end{restatable}
\vspace{-3mm}
The result is easy to parse: to approximate the output at a specific timestamp $k$, the width bound is similar to that of an MLP $(v_i)_{i=1}^k\overset{g_k}{\mapsto} \hat y_k$, which would be $4 r^2 C_{T_k}^2 S^3\epsilon^{-2}$~(see Thm.~\ref{thm:barron_thm}). The additional factor $\|\Omega_k\|^2_2$ appears because $g$ operates on the hidden state and not from the input. Hence, the MLP has to pay a \textbf{penalty for reconstruction}. The result is based on the computation of the Barron constant for the map $f_k$, which is upper bounded by $\|\Omega_k\|_2C_{T_k}$~(see appendix). In the next paragraph, we study how complexity is affected by the additional requirement that the map $g$ should be able to interpolate in-between $T_k$s, based on the first coordinate in the hidden state $x_{k,1}=k$.

\vspace{-4mm}
\paragraph{Step 2: arbitrary timestamp.} To construct a time-independent global approximation, we use recent results from harmonic analysis~\citep{tlas2022bump} to construct a proper domain relaxation and operate with a single MLP. This computation leads to a formula for the Barron complexity for function sequences, where an input dimension determines the function the MLP should implement. In addition to the definition $C_f = \int_{\R^n} \|\omega\|_2 |\mathcal{F} (\omega)|d\omega$, define the Fourier norm $C_f' = \int_{\R^n} |\mathcal{F} (\omega)|d\omega$. 

\begin{restatable}[Combination of Barron Maps]{theorem}{mainbarronk}
\label{thm:comb_barron}
Let $f:[1, 2, \dots, L] \times \R^n\to\R$ be such that each $f(k, \cdot)$ is Barron with constant $C_{f_k}$ and Fourier norm $C_{f_k}'$. There exists an extension $\tilde f:\R^{n+1}\to\R$ of $f$, s.t. $\tilde f$ is Barron with constant $\tilde C \le \mathcal{O}(\sum_{k=1}^L C_{f_k} + C_{f_k}') = \mathcal{O}(L)$ and $\tilde f(x,k) = f(x,k), \forall x\in\R^n$ and $k\in 1, 2, \dots, L$.
\end{restatable}
This result, which essentially confirms the need, under no further assumptions on $T$, to blow up the number of hidden neurons with the sequence length~(unless some smoothness is assumed, see Remark~\ref{rmk:compression}), directly leads to our main result~(Thm.~\ref{thm:universal}) under the assumption that the RNN has perfect memory, guaranteed if $N\ge\dim(\mathcal{V})$~(Thm.~\ref{thm:perfect}).

\textit{Proof of Thm.~\ref{thm:universal}.} Let $f_k := T_k\circ \Omega_k:\R^{2N}\to\R^S$. By Prop.~\ref{prop:single_timestamp}, $C_{f_k}=\|\Omega_k C_{T_k}\|$. $C_{f_k}'$ scales similarly. We have $f_k(x_k) = T_k(v_{1:k})$ thanks to Thm.~\ref{thm:perfect}. We then apply Thm.~\ref{thm:comb_barron} to the sequence of $f_k$~(one for each timestamp): there exists an interpolating function $\tilde f$~(time as first component followed by the $2N$ arguments of $f_k$) with Barron constant $\mathcal{O}(\sum_{k=1}^L C_{f_k} + C_{f_k}') = \mathcal{O}(L)$. We let this be the function the 1-HL MLP $g$ has to approximate, and apply Thm.~\ref{thm:barron_thm}.

\section{Validation and Discussion}
\label{sec:discussion}
In this section, we revisit and validate our claims, and further discuss the practical insights associated with input reconstruction from the linear RNN hidden state.

\subsection{Conditioning and role of complex numbers}
\label{sec:exp_rec_vander}

\begin{figure}[ht]
    \centering
    \vspace{-2mm}
    \includegraphics[width=0.95\linewidth]{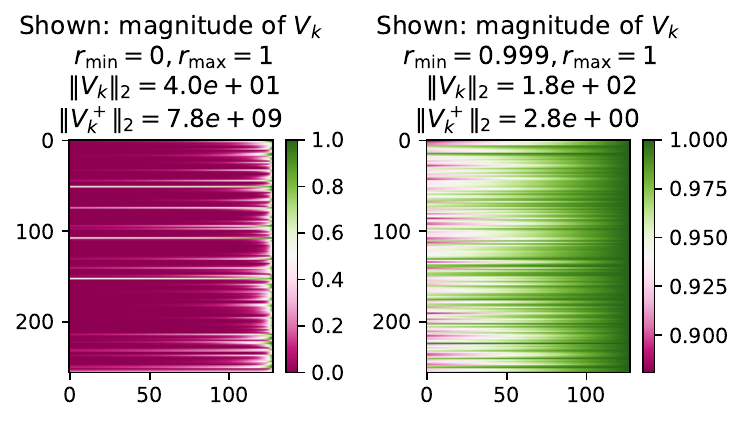}
    \vspace{-4mm}
    \caption{Effect of eigenvalue magnitude on conditioning. $L = 128$, $N = 2L =256$, $\lambda_i\sim \mathbb{T}[r_{\min}, r_{\max}]$.}
    \label{fig:vander_study}
    \vspace{-4mm}
\end{figure}

\begin{figure*}[h]
    \centering
    \includegraphics[width=0.93\textwidth,trim=0pt 0pt -10pt 0pt, clip]{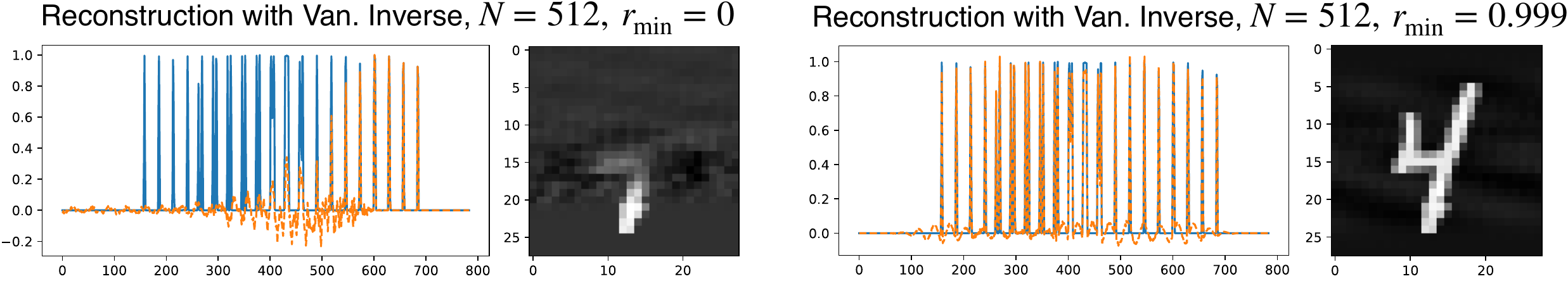}
    \vspace{-4mm}
    \caption{Reconstruction of MNIST digits from the final linear RNN hidden state using the Vandermonde inverse. For $r_{\min}=0$, the Vandermonde is ill-conditioned~(Fig.~\ref{fig:vander_study}) and hence only the recent past can be reconstructed. For $r_{min}=0.99$ we can reconstruct the whole image. See Fig.~\ref{fig:MNIST_PF_rec_MLP} for results on learned reconstructions.}
    \label{fig:MNIST_rec_random_van_main}
    \vspace{-2mm}
\end{figure*}

In \S\ref{sec:main_idea}, we discussed the main idea behind reconstruction from the linear RNN hidden state. Let us restrict attention to the last timestamp: under no sparseness assumption~(discussed at the end of this subsection) $x_L = V_k \bar v^\top$, with $V_L\in\C^{N\times L}$, hence $\bar v^\top = V_L^+x_L$, where $V_L^+ = (V_L^\top V_L)^{-1}V_L^\top =:\Omega_L$. Indeed, if the eigenvalues $(\lambda_i)_{i=1}^N$ are distinct and $N\ge L$, $V_L^\top V_L\in\C^{L\times L}$ is invertible since the Vandermonde determinant formula ensures $V_L$ has full column rank\footnote{By contradiction, assume $V_L^\top V_L$ has the zero eigenvalue. This implies there exists $x$ such that $V_L^\top V_L x = 0$, hence $x^\top V_L^\top V_L x = \|V_L x\| = 0$, which is true if and only if $V_L x=0$, meaning $V_L$ is not full column rank.}. Such computation does not take into account a \textit{numerical issue}: if $V_L$ is \textbf{ill-conditioned}, $\|V_L^+\|_2\to\infty$, preventing successful reconstruction. Instead, when initializing or learning the RNN, not only can we assume the eigenvalues are distinct, but we can also control them. Let us pick $\lambda_i s$ uniformly on the complex ring in between $[r_{\min},r_{\max}]\subseteq[0,1]$: For $i\in[N]$, we pick $\lambda_i$ i.i.d with
\begin{equation*}
    \lambda_i\sim \mathbb{T}[r_{\min}, r_{\max}] := \{\lambda\in\C \ | \ r_{\min}\le|\lambda|\le r_{\max}\}.
\end{equation*}
Initialization of $\Lambda$ \textbf{close to the unit circle} is often used in the context of modern state-space models to unlock long-range reasoning~\citep{orvieto2023resurrecting}. Our theory gives grounding to this strategy: we see in Fig.~\ref{fig:vander_study} that if $r_{\min}\to 1$, then the Vandermonde becomes well conditioned and reconstruction becomes successful~(Fig.~\ref{fig:MNIST_rec_random_van_main} + appendix).

One can also link this effect to vanishing gradients: if $|\lambda_i|\to0$ then the hidden state ``forgets'' inputs at a rate $\lambda^k$. This is the reason why in Fig.~\ref{fig:MNIST_rec_random_van_main} we observe successful reconstruction only for the recent past if $r_{\min}=0$.

\vspace{-3mm}
\paragraph{Crucial role of complex numbers.} If $\Lambda$ is real the condition number of $V_k$ grows exponentially with $N$~\citep{gautschi1987lower}. In the complex setting it is possible to make the condition number of $V_k$ exactly 1 by e.g. choosing the $N$th-roots of unity as eigenvalues of $\Lambda$~\citep{gautschi1975optimally,cordova1990vandermonde}. This fact provides a \textit{precise justification for the use of complex numbers in the recurrent computation}: diagonal real recurrent RNNs can be also implemented fast with parallel scans~\citep{smith2023simplified}, but suffer from an inherent information loss.

\vspace{-3mm}
\paragraph{Comparison with FFT.} The case of equally-spaced eigenvalues on the disk coincides with the computation of the Fourier Transform. Perfect reconstruction in this setting is guaranteed by the inverse Fourier Transform Theorem~\citep{folland2009fourier}. While this would motivate the use of unitary RNNs~\citep{arjovsky2016unitary, helfrich2018orthogonal}, as we will see in \S\ref{sec:odes} best performing learned architectures do not have this property: hidden-state features with vanishing memory are often the best option. 

\vspace{-4mm}
\paragraph{Sparseness improves conditioning.} In the last section, we discussed how input sparseness~(on a basis of $P$ functions) results in a reduced hidden state dimension requirement. Specifically, if $\dim(\mathcal{V})=P$, i.e. there exists $\Psi\in\R^{L\times P}$ such that for any $\bar v\in\mathcal{V}$ there exists $\alpha^v\in\R^P$ such that $\bar v^\top = \Psi \alpha^v$, then an RNN with width $N=P$ achieves perfect recall for inputs in $\mathcal{V}$. In Fig.~\ref{fig:effect_P}~(appendix) we show that sparseness also improves conditioning of the reconstruction map $\Omega_L$, appearing in Prop.~\ref{prop:single_timestamp}.

\subsection{Learned architecture for reconstruction}
\label{sec:learned_rec}
In the previous subsection we studied the reconstruction map computed from the parameters of a \textbf{random} linear RNN. While our theory provides satisfactory approximation guarantees in this setting, performance is much stronger if we allow the linear RNN to be \textit{\textbf{trained} jointly with the reconstruction map}. We use here a simple LRU block~(at inference time, a linear RNN~\citep{orvieto2023resurrecting}) followed by a linear reconstruction head $\Omega$. This framework, while perfectly in line with our settings in this paper, allows the RNN eigenvalues and the decoder to synchronously adapt to the specific data structure. Out of curiosity, we also test here the performance of non-linear reconstruction from the RNN hidden state with a 1-HL MLP with $2048$ hidden units\footnote{This is not the function $g$ that allows approximation of $T_k$. Here the MLP serves as non-linear decoder.}. We use data from the long-range arena~\citep{tay2020long}, specifically sequential MNIST~(784 tokens) and PathFinder~(1024 tokens) and train our model\footnote{We train on a single Nvidia RTX A5000 for 100 epochs.} under the usual guidelines from random initialization~(see discussion in~\cite{orvieto2023resurrecting}). 
\vspace{-3mm}
\begin{enumerate}[left=0.1em, itemsep=1pt]
    \item For both datasets, the reconstruction error~(input of reconstruction map is the last hidden state) vanishes as the RNN hidden dimension increases. At $N=256$, RNN width used in practice on the long-range arena~\citep{gu2021efficiently, orvieto2023resurrecting}, reconstruction is nearly perfect. For $N=1024$, the error is negligible~(as predicted by Thm.~\ref{thm:perfect}) since $N\ge L$.
    \item Reconstruction is surprisingly accurate at small values of $N$ if the decoder is non-linear. Instead, for $N=1024$, the error is smallest with a linear decoder, a finding we believe is rooted in optimization of wide MLPs. 
\end{enumerate}


\subsection{Approximation of non-linear ODEs.}
\label{sec:odes}
\begin{figure}[h]
\vspace{2mm}
    \centering
    \includegraphics[width=0.99\linewidth]{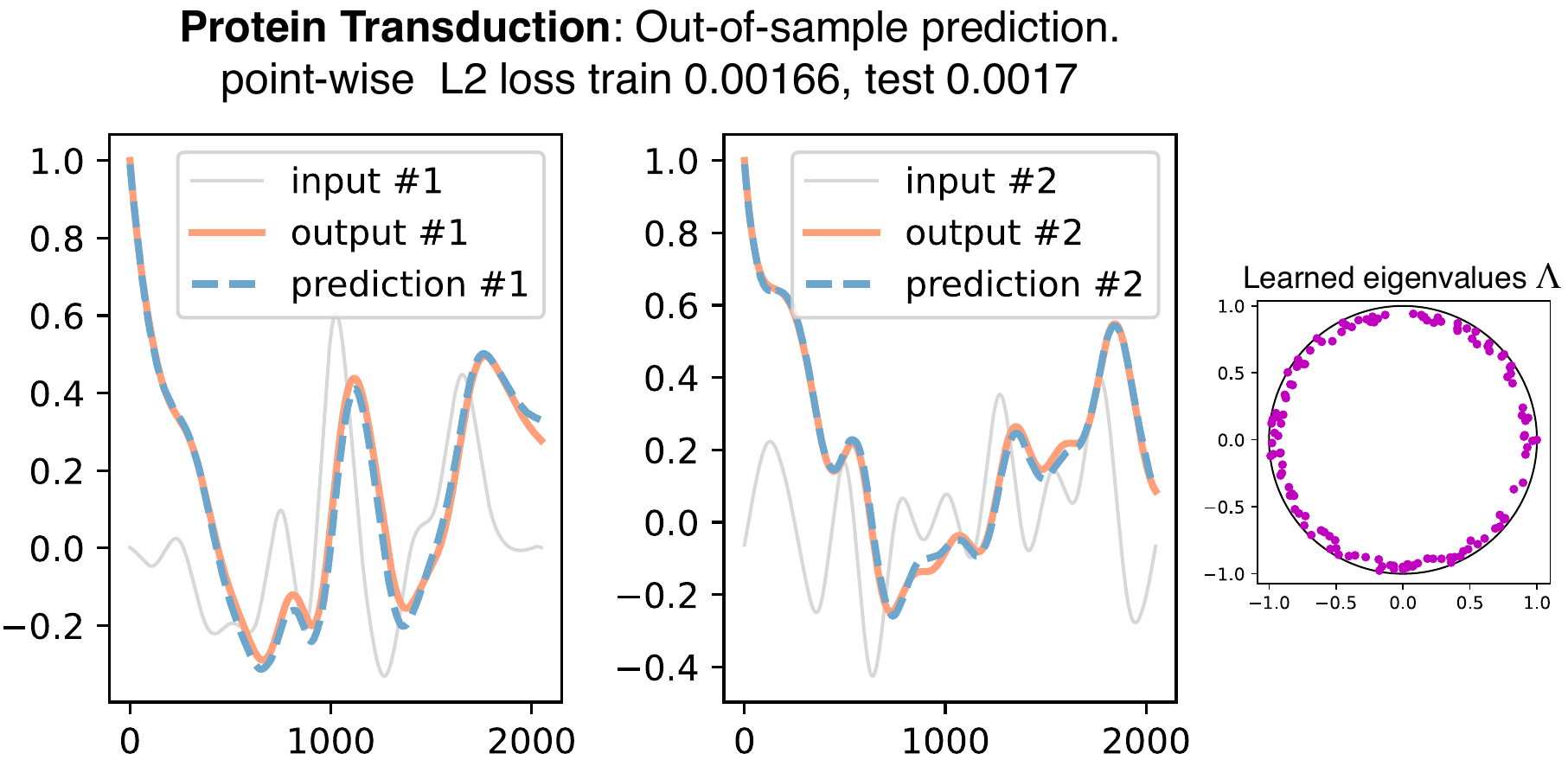}
    \vspace{-2mm}
    \caption{Trained linear RNN + MLP architecture~(1 layer) learns a non-linear sequence-to-sequence map. Additional experiments can be found in the appendix.}
    \label{fig:ptpz}
    \vspace{-4mm}
\end{figure}

In \S\ref{sec:exp_rec_vander}\&\ref{sec:learned_rec} we studied the information content of linear RNNs hidden states and verified empirically that near-perfect input reconstruction is possible, even if $N<L$. We now conclude the paper with results verifying our main claim~(Thm.~\ref{thm:universal}): a single MLP applied to the RNN hidden states, independent of the timestamp, is able to reconstruct the output of regular sequence-to-sequence maps~(Def.~\ref{def:seq}). An example of a sequence-to-sequence map is the flow of a controlled differential equation, of form $\dot z_t = f(z_t, v_t), y_t = h(z_t)$, where $(v_t)_{t}$ is the input, $f$ is a non-linear multidimensional function, and $h$ projects the multidimensional state $z_t$ into a one-dimensional output. Such systems can model complex interactions and stand at the root of physics. We limit our discussion here to the problem of learning a \textit{protein transduction~(PT) system}~\citep{vyshemirsky2008bayesian}.\textit{The reader can find results for other non-linear ODEs in the appendix}. \\
In PT, $z_t$ is 5-dimensional, and $f$ includes multiplicative interactions between components of $z_t$ as well as feedback~(see equations in the appendix). We include an input to the right-hand-side of the first ODE while integrating with Runge-Kutta 45, and take $h$ as the projection of the first component of $z_t$. We learn to mimic the PT system with a linear RNN with $N=128$, followed by a 1-HL MLP with $D=256$. We sample $10k$ smooth inputs and train on the simulated outputs for $L=2048$ integration steps. We test on $1k$ additional trajectories. The results in Fig.~\ref{fig:ptpz} clearly show that the architecture studied in this paper is able to learn the PT sequence-to-sequence map: plotted are two out-of-sample trajectories~(not used during training), showcasing that our model prediction matches the PT dynamics up to a small error~(as confirmed by the low test loss). We also include a depiction of the learned eigenvalues, to confirm that the RNN has indeed vanishing memory -- but this does not necessarily imply information loss.

\vspace{-1mm}
\section{Conclusion}
In this paper we studied the expressive power of architectures combining linear RNNs with position-wise MLPs. These models recently exhibited remarkable performance in the context of long-range reasoning. In our theoretical framework, these two blocks work in symbiosis while having distinct roles: the linear RNN provides a compressed~(where possible) representation of the input sequence, while the MLP performs non-linear processing. We believe our \textit{separation of concerns} principle takes us one step further in understanding how to design and conceptualize deep state-space models: non-linear sequential processing can be performed by a time-independent component~(the MLP). Further, the use of complex numbers in the recurrence unlocks lossless compression via a well-conditioned reconstruction map. In addition, our analysis provides guidelines for asymptotic scaling of architecture components, and can lead to novel initialization strategies based on local objectives~(e.g. successful reconstruction). 
\vspace{-3mm}

\paragraph{Future work.} Our work leaves open interesting avenues for future research, e.g. the finite-alphabet setting, the finite precision setting, and the framework of Turing completeness. Regarding the expressivity of recent selective RNNs such as Mamba and Griffin, an extension of our theory can be found in~\citet{cirone2024theoretical}. Last, we note that, as remarked in \S\ref{sec:hippo}, our framework does not model errors in reconstruction present at moderate hidden state dimensions. We believe a unified formal analysis, comprising assumptions such as smoothness in the inputs and tools such as the~\citet{johnson1986extensions} lemma, can fill this gap.
\vspace{-3mm}

\paragraph{Insights from Mamba \& \textit{Never train from scratch}.} The first version of this paper, in the form of a short note, dates back to May 2023. In late 2023, selective SSMs~\citep{gu2023mamba} were introduced: these SSMs \textit{do not use of complex numbers}, have input-controlled state transitions, and are almost exclusively trained autoregressively~(on text). \citet{amos2023never} brilliantly pointed out that next-token prediction leads to a conceptually distinct objective compared to standard supervised learning: while training from scratch requires careful parametrization, initialization, and memorization, pretraining on denoising objectives can be more forgiving: transformers and less sophisticated S4 variants are able to get high accuracy on the LRA benchmark~\citep{tay2020long}. Our paper, and specifically our discussion on complex numbers, is instead motivated by interesting results and the ablations holding for non-selective SSMs trained from scratch~\citep{gu2021efficiently,orvieto2023resurrecting}. We believe the supervised learning setting is scientifically interesting and relevant, and that the analysis presented here provides a precise motivation for effective design strategies. While the next-token prediction paradigm is becoming a standard, the peculiarities of supervised learning can yield insights into the next generation of generative models.

\section*{Acknowledgements}
Antonio Orvieto acknowledges the financial support of the Hector Foundation.

\section*{Impact Statement}
The authors do not foresee any societal consequences or ethical concerns arising from their work. This is a theoretical paper.

\bibliography{main}
\bibliographystyle{icml2024}

\appendix
\onecolumn
\begin{center}
{\Large \textbf{Appendix}}
\end{center}

\section{Related Works}

\paragraph{Issues with attention for long-range reasoning.} Efficient processing of long sequences is an important open question in deep learning. Attention-based transformers~\citep{vaswani2017attention} provide a scalable approach but suffer from \textit{quadratically increasing complexity in inference/memory} as the sequence length grows. While many approaches exist to alleviate this issue, e.g. efficient memory management~\citep{dao2022flashattention,dao2023flashattention} and architectural modifications~\citep{wang2020linformer, kitaev2020reformer, child2019generating, beltagy2020longformer}, the sequence length in large language models is usually kept to $2k/4k$ tokens for this reason~(e.g. Llama2~\citep{touvron2023llama}). In addition, in some long-range reasoning tasks~\citep{tay2020long} attention does not seem to provide the correct \textit{inductive bias}, leading to poor performance in addition to high computational costs.
\vspace{-3mm}
\paragraph{Success of modern recurrent layers.} Due to the issues outlined above, the community has witnessed in the last year the rise of new, drastically innovative, \textit{recurrent} alternatives to the attention mechanism, named state-space models~(SSMs). The first SSM, S4, was introduced by \cite{gu2021efficiently}, and since then, a plethora of variants have been proposed:  LiquidS4~\citep{hasani2022liquid}, DSS~\citep{gupta2022diagonal},
S4D~\citep{gu2022parameterization}, S5~\citep{smith2022simplified}, RWKV~\citep{peng2023rwkv} and RetNet~\citep{sun2023retentive} among others. These models achieve remarkable performance, surpassing all modern attention-based transformer variants by an average $20\%$ accuracy on challenging sequence classification tasks~\citep{tay2020long}. SSMs have reached outstanding results in various domains beyond toy datasets~\citep{nguyen2022s4nd,goel2022sashimi,gu2021efficiently,lu2023structured,zucchet2023online}. SSMs also were successfully applied to language modeling, and are sometimes used in combination with attention~\citep{fu2023hungry, wang2023pretraining, ma2022mega}. At inference time, all SSMs coincide with a stack of linear Recurrent Neural Networks, interleaved with MLPs and normalization. Linearity of the RNNs allows fast parallel processing with FFTs~\citep{gu2022parameterization} or parallel scans~\citep{smith2023simplified}.  
\vspace{-3mm}

\paragraph{The linear recurrent unit~(LRU).} Among modern architectures for long-range reasoning based on recurrent modules, the simplest is perhaps Linear Recurrent Unit~(LRU)~\citep{orvieto2023resurrecting}: while SSMs rely on the discretization of a structured continuous-time latent dynamical system, the LRU is directly designed for discrete-time systems~(token sequences), and combines easy hyperparameter tuning with solid performance and scalability. The only differences between the LRU and the standard RNN update $x_{k} = A x_{k-1} + B u_k$~($u$ is the input at a specific layer and $x$ is the hidden-state, then fed into a position-wise MLP) are (1) the system operates in the complex domain~(required for expressivity, see discussion in~\cite{orvieto2023resurrecting})~(2) to enhance stability and better control how quickly gradients vanish, $A$~(diagonal) is learned using polar parametrization and log-transformed magnitude and phase. Finally, (3) the recurrence is normalized through an extra optimizable parameter that scales the input to stabilize signal propagation. The parametrization of linear RNNs of~\citep{orvieto2023resurrecting} was found to be effective also in surpassing deep LSTMs and GRUs~\citep{zucchet2023gated}. We use the LRU codebase\footnote{\url{https://github.com/NicolasZucchet/minimal-LRU/tree/main/lru}} as a starting point for our experiments, when the linear RNN is learned.
\vspace{-3mm}
\paragraph{Approximation theory for MLP and non-linear RNNs.} The approximation properties of deep neural networks with ReLU activations are well studied. While recent advances concern the effect of depth~\citep{lu2017expressive}, the study by \citet{pinkus1999approximation}, as well as previous works~\citep{funahashi1989approximate,hornik1989multilayer,hornik1991approximation, barron1993universal}, already established the power of neural networks with a single hidden layer, which can approximate arbitrary continuous non-linear maps on compacts as the size of the hidden layer grows to infinity. The cleanest result is perhaps the one of \citet{barron1993universal}, that we heavily use in this paper.\\
In the context of non-linear RNN approximation of dynamical systems~(e.g. in neuroscience), the state-to-state computation can be seen as part of an MLP~(see e.g.~\citet{hanson2020universal}): we have $x_k = \sigma(A x_{k-1} + B u_k)$, where $\sigma$ is is a non-linearity. As a result, wide non-linear RNNs can in principle approximate non-linear dyamical systems, as we discuss in detail in App.~\ref{app:rw_MLP}.\\
Meanwhile, linear RNNs, where $x_k = A x_{k-1} + B u_k$, have often been considered of minor interest, as they equivalent in approximation power to convolutions~\citep{li2022approximation}~(see App.~\ref{app:rw_MLP}). In this paper we take a different approach: we show that when sufficiently wide, the linear RNNs \emph{do not} form a bottleneck, and the architecture maintains universality through the application of the pointwise MLP on the hidden state, as done in recent SSMs achieving state-of-the-art results~\citep{gu2021efficiently, smith2022simplified, orvieto2023resurrecting}. As motivated thoroughly in the paper, this architecture unlocks parallel computation, in contrast to what is possible with directly placing non-linearities in the recurrence.

\section{Approximation theory for (non-linear) RNNs}
\label{app:rw_MLP}

We recall a result on universality of MLPs already stated in the main paper.

\barron*
\barronthm*

\subsection{Guarantees for RNNs with recurrent non-linearities} Research on universality of non-linear RNNs dates back to~\citep{siegelmann1992computational}. We present here a more recent result by \citet{hanson2020universal}.

\begin{theorem}[Universality of non-linear RNNs] Consider the continuous-time non-linear dynamical system with form
\begin{equation}
    \dot{\bar{x}}(t) = \bar f(\bar x(t), u(t)),\quad \bar y(t) = h(\bar x(t)),
    \label{eq:non-linear-RNN-ct}
\end{equation}
with $\bar x(t)\in\R^{\bar N}$, $u(t)\in\R^M$. Under some technical assumptions~(bounded input, non-chaotic $f$), for any $\epsilon>0$ there exists a non-linear RNN
\begin{equation}
    \dot x(t) = -\frac{1}{\tau} x(t) + \sigma(A x(t) + Bu(t)), \quad y(t) = Cx(t),
    \label{eq:non-linear-RNNapprox-ct}
\end{equation}
for some non-polynomial $\sigma$, $\tau>0$, $A\in\R^{N\times N}$, $B\in\R^{N\times M}$, $C\in\R^{M\times N}$ that approximates the solution to Eq.~\ref{eq:non-linear-RNN-ct} up to error $\epsilon$ uniformly in time, on compact sets of inputs.
\end{theorem}
The result above typically involves taking $N$~(RNN width) to infinity.

\textit{Proof.} We briefly outline the idea behind the proof, and invite the reader to refer to~\cite{hanson2020universal} for details. Approximating the solution to Eq.~~\ref{eq:non-linear-RNN-ct} is equivalent to approximating the infinitesimal solution generator, which is a non-linear function of $(x, u)$. By Barron's Theorem~(Thm.~\ref{thm:barron_thm}), this generator can be approximated by a one-layer MLP, that is exactly the right-hand side of Eq.~\ref{eq:non-linear-RNNapprox-ct}.
\proofend

\subsection{Guarantees for linear RNNs} Simply taking out the non-linearity from the recurrence in Eq.~\ref{eq:non-linear-RNNapprox-ct} restricts the function class to convolutions. To start, recall that the linear continuous-time RNN on one-dimensional inputs
\begin{equation*}
    \dot x(t) = A x(t) + B u(t), \quad y(t) = Cx(t),
\end{equation*}
with $A\in\R^{N\times N}$, $B \in\R^{N\times M}$, $C \in\R^{1\times N}$
has solutions given by a convolution.
\begin{equation*}
    x(t) = \int_0^t C^\top e^{As}B u(t-s) ds =: \int_0^t \rho(s)^\top u(t-s) ds =: \sum_i(\rho^i\star u^i)_t.
\end{equation*}
Let us call $\hat{\mathcal{H}_N}$ the class of functionals parametrizable with linear RNNs with hidden state of dimension $N$, and $\hat{\mathcal{H}} = \cup_{N\in\mathbb{N}_+}\mathcal{H}_N$. 
\begin{equation*}
    \mathcal{H}_N :=\left\{ \{H_t
: t \in \R\}, H_t(u) = \int_{0}^{t}C^\top e^{As} Bu(t-s) ds, C\in\R^{1\times N}, A\in\R^{N\times N}, B\in \R^{N\times M}\right\}.
\end{equation*}
It turns out that convolutions are dense in the class of linear functionals. Let $\mathcal{U} = C_0(\R,\R^d)$ with norm $\|u\| = \sup_{t\in\R} \|u(t)\|_\infty$.
\begin{theorem}[Linear functionals in $C_0(\R,\R^d)$ are convolutions \citep{li2022approximation}]
Let $\{H_t
: t \in \R\}$ be a
family of continuous, linear, causal, regular, and time-homogeneous functionals on $\mathcal{U}$, i.e. such that
\begin{enumerate}
    \item (Continuous)  $\forall t\in\R$, $\sup_{\|u\|<1} H_t(u)<\infty$.
    \item (Linear) $\forall t\in\R$, $u,v\in \mathcal{U}$ and $\nu,\lambda\in \R$, we have $H_t(\lambda u + \nu v) = \lambda H_t(u) + \nu H_t(v)$.
    \item (Causal) For all $u,v\in\mathcal{U}$ such that $u(s)=v(s)$ for all $s\le t$, he have $H_t(v)=H_t(u)$.
    \item (Regular) Let $(u^n)$ be a sequence in $\mathcal{U}$ s.t. $u^n(s)\to 0$ for almost every $s\in\R$, then, for all $t\in\R$, we have $\lim_{n\to\infty} H_t(u^n)=0$.
    \item (Time Homogeneous) For all $u\in \mathcal{U}$ let $u^{\tau}_t = u(t-\tau)$, then $H_{t}(u^{\tau}) = H_{t+\tau}(u)$. 
\end{enumerate}
Then, for any $\{H_t
: t \in \R\}$ there exist a function (a kernel) $\bar \rho:\R_+\to\R^M$ such that for all $t\in\R$.
\begin{equation*}
    H_t(u) = \int_{0}^\infty \rho(s)^{\top} u(t-s) ds = \sum_i(\bar \rho^i\star u^i)_t.
\end{equation*}
\end{theorem}

\begin{theorem}[Linear RNNs can parametrize any convolution \citep{li2022approximation}]
Let $\{H_t
: t \in \R\}$ be a
family of continuous, linear, causal, regular, and time-homogeneous functionals on $\mathcal{U}$. Then,
for any $\epsilon > 0$ there exists $\{\hat H_t: t \in \R\}\in\hat{\mathcal{H}}$ such that
$$
    \sup_{t\in\R}\sup_{\|u\|<1} \|H_t(u)-\hat H_t(u)\|\le \epsilon.
$$
\label{thm:linear_rnn}
\end{theorem}
The result above typically involves taking $N$~(RNN width) to infinity.

\paragraph{This paper.} There is a sizable gap between the power of nonlinear and linear RNNs. We show in this paper that placing a nonlinearity at the output of linear RNNs~(unlocking parallel computation) allows approximation of arbitrary regular non-linear sequence-to-sequence mappings.

\newpage
\section{Details on Multidimensional Input Reconstruction}

In the main text, we showed that linear diagonal RNN computations on one-dimensional input sequences can be written in matrix form using a Vandermonde matrix~(Sec.~\ref{sec:vandermonde}). For convenience of the reader, we repeat the reasoning here: let $H=M=1$, the encoder $e$ be the identity, and $B = (1,1,\dots, 1)^\top$. Then, eq.~(\ref{eq:lin_rnn_unroll}) can be written as
\begin{equation*}
    x_k   =
    \begin{pmatrix}
    \lambda_1^{k-1}&\lambda_1^{k-2} &\cdots& \lambda_1&1\\
    \lambda_2^{k-1}&\lambda_2^{k-2} &\cdots& \lambda_2&1\\
    \vdots&\vdots &\ddots&\vdots&\vdots\\
    \lambda_N^{k-1}&\lambda_N^{k-2} &\cdots& \lambda_N&1\\
    \end{pmatrix}
    \begin{pmatrix}
    u_1 \\ u_{2} \\ \vdots \\ u_k
    \end{pmatrix}= V_k u_{1:k}^\top.
\end{equation*}
where $u_{1:k}= v_{1:k} = (v_i)_{i=1}^k \in\R^{1\times k}$, and $V_k$ is a Vandermonde matrix. As long as $N\ge k$, we can hope to recover $u_{1:k}$ by pseudoinversion of the Vandermonde:
\begin{equation*}
    v_{1:k}^\top = u_{1:k}^\top = V_k^+ x_k,
\end{equation*}

Here, we give  details on the design of input projections such that the RNN output from multidimensional inputs can also be seen as matrix multiplication.  Let us define
\begin{equation*}
    \vect(v_{1:k}) := \begin{pmatrix}
v_{1,1:k}^\top \vspace{1mm}\\
 v_{2,1:k}^\top \\
 \vdots \\
 v_{M,1:k}^\top
    \end{pmatrix}\in\R^{kM},
\end{equation*}

The matrix $B$ we are going to use in our linear diagonal RNN is
\begin{equation*}
    B = \begin{pmatrix}
 1_{N'\times 1} & \cdots & 0_{N'\times 1} & 0_{N'\times 1}\\
 0_{N'\times 1} & \cdots & 0_{N'\times 1} & 0_{N'\times 1}\\
  \vdots & \ddots & \vdots & \vdots\\
 0_{N'\times 1} & \cdots & 1_{N'\times 1} & 0_{N'\times 1}\\
 0_{N'\times 1} & \cdots & 0_{N'\times 1} & 1_{N'\times 1}\\
\end{pmatrix},
\end{equation*}
where we select $N = MN'$. With this choice, the linear diagonal RNN output can be written as
\begin{equation*}
    x_k  =
    \begin{pmatrix}
V_{k,1} & \rvline & & \rvline & &  \rvline & \\
\hline
  &\rvline & V_{k,2}& \rvline & &  \rvline & \\
\hline
  &\rvline & & \rvline & \ddots &  \rvline & \\
\hline
  &\rvline & & \rvline && \rvline  &  V_{k,M} \\
\end{pmatrix}
    \begin{pmatrix}
v_{1,1:k}^\top \\
    \vspace{1mm}
 v_{2,1:k}^\top \\
 \vdots \\
 v_{M,1:k}^\top
    \end{pmatrix} = \tilde V_k  \begin{pmatrix}1\\
    \vect(v_{:,1:k})\end{pmatrix},
\end{equation*}

where $V_k = \diag(V_{k,1},V_{k,2},\cdots,V_{k,M})$, and $V_{k,j}\in\C^{N'\times k}$ is the Vandermonde matrix corresponding to the block $\Lambda_j$ in the diagonal recurrent matrix $\Lambda$:
\begin{align*}
&\Lambda = \diag( \Lambda_1,\Lambda_2,\cdots,\Lambda_M)\in\C^{(N'M)\times(N'M)},\\
    &\Lambda_j = \diag(\lambda_{1,j},\lambda_{2,j},\cdots,\lambda_{N',j})\in\C^{N'\times N'},\\
    &V_{k,j} = \begin{pmatrix}
    \lambda_{1,j}^{k-1}&\lambda_{1,j}^{k-2} &\cdots& \lambda_{1,j}&1\\
    \lambda_{2,j}^{k-1}&\lambda_{2,j}^{k-2} &\cdots& \lambda_{2,j}&1\\
    \vdots&\vdots &\ddots&\vdots&\vdots\\
    \lambda_{N',j}^{k-1}&\lambda_{N',j}^{k-2} &\cdots& \lambda_{N',j}&1\\
    \end{pmatrix}\in\C^{N'\times k}.
\end{align*}
This construction effectively decouples each input dimension and reduces the discussion to the one-dimensional setting: invertibility of $V_k$ is guaranteed by invertibility of each block, provided $N'\ge L$ and that eigenvalues are distinct. Slight changes can be made to keep also track of the timestamp~(see Sec.~\ref{sec:main_idea}) and to adapt to the sparse setting~(see Sec.~\ref{sec:sparse}).

\newpage
\section{Expressivity Proofs}

One of our main steps involved computation of the Barron constant of the function mapping the RNN hidden state to the output.
\mainMLPsing*

Since $\Omega_k$ is a matrix, the result can be proved by computing the Barron constant of a function where the argument is the output of a linear map.

\begin{restatable}[Change of variables]{lemma}{cov}
Let $A\in\R^{p\times n}$ and $f(x) = g(A x)$, then 
$$C_f = \|A\|_2 C_g.$$
\end{restatable}

\textit{Proof.} The inverse Fourier transform formula directly leads to
\begin{equation}
    f(x) = \int_{\R^p} e^{i \langle p, A x\rangle} \mathcal{G}(\xi)d\xi.
\end{equation}
Let us now compute the Fourier Transform of $f$.
\begin{align}
    \F(\omega) &= \int_{\R^n} e^{-i\langle \omega, x \rangle} f(x) dx\\
    &= \int_{\R^n} e^{-i\langle \omega, x \rangle} \left[\int_{\R^p} e^{i \langle \xi, A x\rangle} \mathcal{G}(\xi)d\xi\right] dx\\
    &= \int_{\R^n}\int_{\R^p} e^{-i\langle \omega, x \rangle}  e^{i \langle  A^\top \xi, x\rangle} \mathcal{G}(\xi)d\xi dx\\
    &= \int_{\R^p}\left[\int_{\R^n} e^{i \langle A^\top\xi  - \omega, x\rangle}dx\right] \mathcal{G}(\xi)d\xi.
\end{align}

Recall now the definition of the Dirac delta:
\begin{equation}
\delta(z) = \frac{1}{2\pi}\int_{\R}e^{i\nu z}d\nu.
\end{equation}
Therefore
\begin{equation}
    \F(\omega) =\int_{\R^p} \delta(A^\top\xi  - \omega)\mathcal{G}(\xi)d\xi.
\end{equation}
Note that this is a singular measure in $\R^n$: lives in a linear $p$-dimensional subspace. Further, note that
\begin{align}
    C_f &= \int_{\R^n}\|\omega\|_2\cdot|\F(\omega)|d\omega\\
    &= \int_{\R^n}\|\omega\|_2\cdot\big|\int_{\R^p} \delta(A^\top\xi  - \omega)\mathcal{G}(\xi)d\xi\big|d\omega\\
    &\le \int_{\R^n}\|\omega\|_2\cdot\int_{\R^p} \delta(A^\top\xi  - \omega)|\mathcal{G}(\xi)|d\xi d\omega\\
    &\le \int_{\R^p}\left[\int_{\R^n}\|\omega\|_2 \delta(A^\top\xi  - \omega) d\omega\right] |\mathcal{G}(\xi)|d\xi\\
    &= \int_{\R^p}\|A^\top\xi\|_2\cdot  |\mathcal{G}(\xi)|d\xi\\
    &= \|A^\top\|_2\int_{\R^p}\|\xi\|_2\cdot  |\mathcal{G}(\xi)|d\xi.
\end{align}
The proof is done, since $\|A^\top\|_2 = \|A\|$~(same singular values), and $C_g = \int_{\R^p}\|\xi\|_2\cdot  |\mathcal{G}(\xi)|d\xi$.
\proofend

We now proceed proving the complexity for interpolation of sequences of Barron functions. This directly implies our main result~(Thm.\ref{thm:universal}).

\mainbarronk*

\textit{Proof.} Let us consider the following definition for $\tilde f$:
\begin{equation}
    \tilde f(x,t) = \sum_{k=1}^L f(x,k) h(t-k),
\end{equation}
where $h:\R\to\R$ is a filter~(see discussion after the proof) with support in $[-1,1]$. Compactness in the support of $h$ leads to the desired property $\tilde f(x,k) = f(x, t)$, for all $k = 1,2,\dots L$. Let us now compute the Fourier transform of $\tilde f$. Frequencies are of the form $\omega=(w,\nu)$, with $w\in\R^n$, $\nu\in\R$. 
\begin{align}
    \tF(\omega) &= \frac{1}{(2\pi)^{n+1}} \int_\R \int_{\R^n} \tilde f(x,t) e^{-i\langle w,x\rangle} e^{-i\nu t} dx dt\\
    &= \frac{1}{(2\pi)^{n+1}} \int_\R \int_{\R^n} \left[\sum_{k=1}^L f(x,k) h(t-k)\right] e^{-i\langle w,x\rangle} e^{-i\nu t} dx dt\\
    &= \frac{1}{(2\pi)^{n+1}} \sum_{k=1}^L \left[ \int_{\R^n} f(x,k)  e^{-i\langle w,x\rangle}  dx\right]\cdot \left[\int_\R h(t-k) e^{-i\nu t} dt\right]\\
    &= \sum_{k=1}^L \tF_k(w) \mathcal{H}(\nu) e^{i\nu k},
\end{align}
where $\mathcal{H}$ is the Fourier transform of $h$, and the factor $e^{i\nu k}$ comes from the shift $h(\cdot -k)$. All in all, we get
\begin{equation}
    \tF(w,\nu) = \mathcal{H}(\nu) \sum_{k=1}^L \tF_k(w)  e^{i\nu k}.
\end{equation}
Trivially, 
\begin{equation}
    |\tF(w,\nu)| \le |\mathcal{H}(\nu)| \sum_{k=1}^L |\tF_k(w)|.
\end{equation}

Therefore:
\begin{align}
    \tilde C &= \int_{\R^{n+1}} \|\omega\|_2 |\tilde{\mathcal{F}} (\omega)| d\omega\\ &\le \int_{\R^n}\int_\R \|(w,\nu)\|_2 \cdot \left[ |\mathcal{H}(\nu)|\sum_{k=1}^L |\tF_k(w)|\right] dw d\nu\\
&=\sum_{k=1}^L \int_{\R^n}\int_\R \|(w,\nu)\|_2 \cdot |\mathcal{H}(\nu)|\cdot|\tF_k(w)| dw d\nu.
    \end{align}
    
Using the triangle inequality:
\begin{align}
    \tilde C &\le \int_{\R^n}\int_\R (\|w\|_2 + |\nu| )\cdot |\mathcal{H}(\nu)|\cdot|\tF_k(w)| dw d\nu\\ &= \int_{\R^n}\int_\R \|w\|_2\cdot |\mathcal{H}(\nu)|\cdot|\tF_k(w)| dw d\nu + \int_{\R^n}\int_\R |v|\cdot |\mathcal{H}(\nu)|\cdot|\tF_k(w)| dw d\nu\\
    &= C_{f_k} C'_{h}+ C_h C_{f_k}'.
\end{align}
This concludes the proof since~(see next paragraph) $C_h$ and $C'_h$ are problem-independent bounded constants.
\hfill $\square$

The proof of the theorem above concludes stating that $C_h$ and $C'_h$ are problem-independent bounded constants. Recall that, in our proof, $h:\R\to\R$ has compact support in $[-1,1]$. We can design $h$ such that its Fourier transform has fast decay:

\begin{restatable}[\cite{tlas2022bump}]{theorem}{tlas}
For any $\delta \in (0,1)$ and any $c >0$  there is a function $h(t)$ which is $C^\infty$, real, even, nonnegative, supported in $[-1,1]$ and whose Fourier transform $\mathcal{H}(\nu)$ is monotone decreasing for $\nu \geq 0$ and satisfies the following double inequality 
\begin{equation}
      \exp(- (1+\epsilon) c \nu^\delta ) \lesssim  \mathcal{H}(\nu)  \lesssim  \, \, \exp(- (1- \epsilon) c \nu^\delta),
\end{equation}
for any $\epsilon > 0$.
\end{restatable}
This result, rooted in the Beurling-Malliavin multiplier theorem\cite{mashreghi2006beurling}, ensures that we can design $h$ on a compact support with exponentially decaying frequencies. This implies that both integrals
\begin{equation}
    C_h = \int_{\R} |\nu| |\mathcal{H} (\nu)|d\nu,\qquad 
    C_h' = \int_{\R} |\mathcal{H} (\nu)|d\nu
\end{equation}
are bounded and independent on the specific form of the function $f$ we want to approximate.

\newpage
\section{Additional experiments}
\subsection{Reconstruction using the Vandermonde inverse~(support to Section~\ref{sec:vandermonde})} 
\label{app:rec_vandermonde}
We consider linear diagonal RNNs with the $N$ diagonal entries of $\Lambda$ sampled inside the unit disk in $\C$, uniformly in angle in between radii $r_{\min}$ and 1. We consider the hidden state $x_L\in\C^N$ computed after $L$ RNN steps, \textit{i.e. after the sequence is completely processed}. We want to recover the input sequence from the hidden state using the Vandermonde inverse $V_L^+$~(see Sec.~\ref{sec:vandermonde}). If $N\ge L$, since under random initialization the determinant of any set of $L$ columns of $V_L$ is positive, we can in theory achieve perfect reconstruction. In practice, $V_L$ is ill conditioned --- especially if $r_{\min}$ is not close to $1$. This causes some problem in the pseudoinverse computation, which may result in imperfect reconstruction. 

\begin{figure}[ht]
    \centering
    \includegraphics[height=0.14\textwidth]{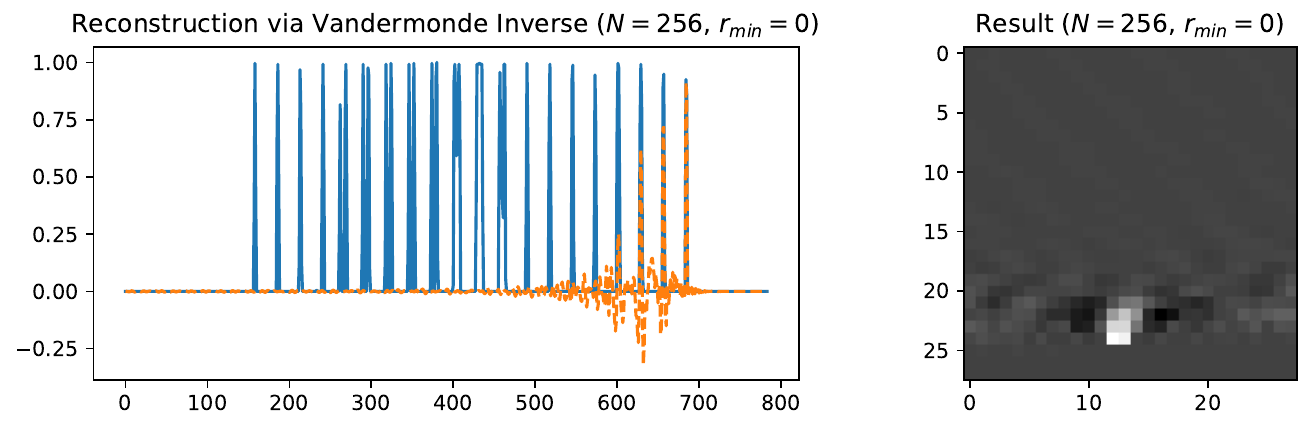}\hspace{10mm}
    \includegraphics[height=0.14\textwidth]{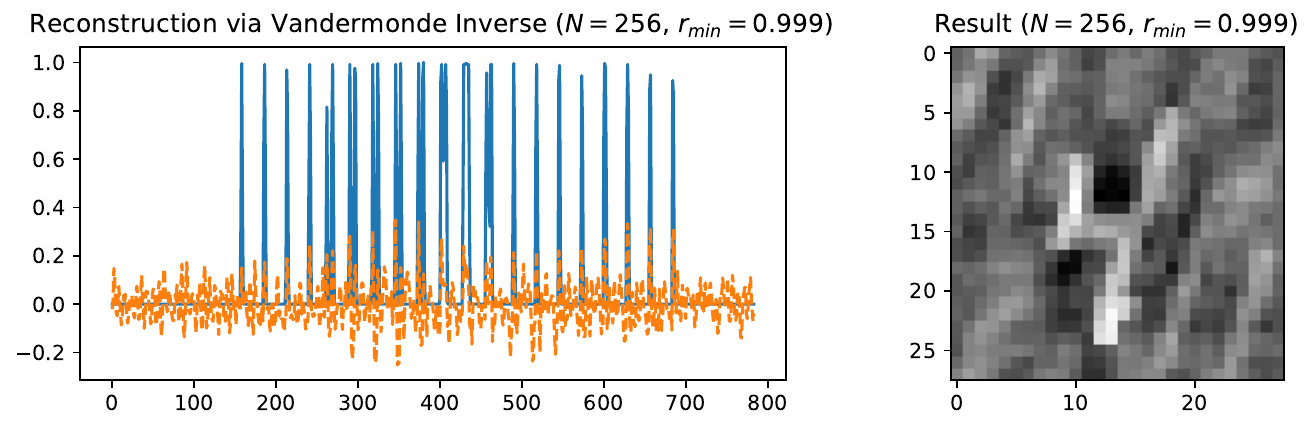}
\\
    \includegraphics[height=0.14\textwidth]{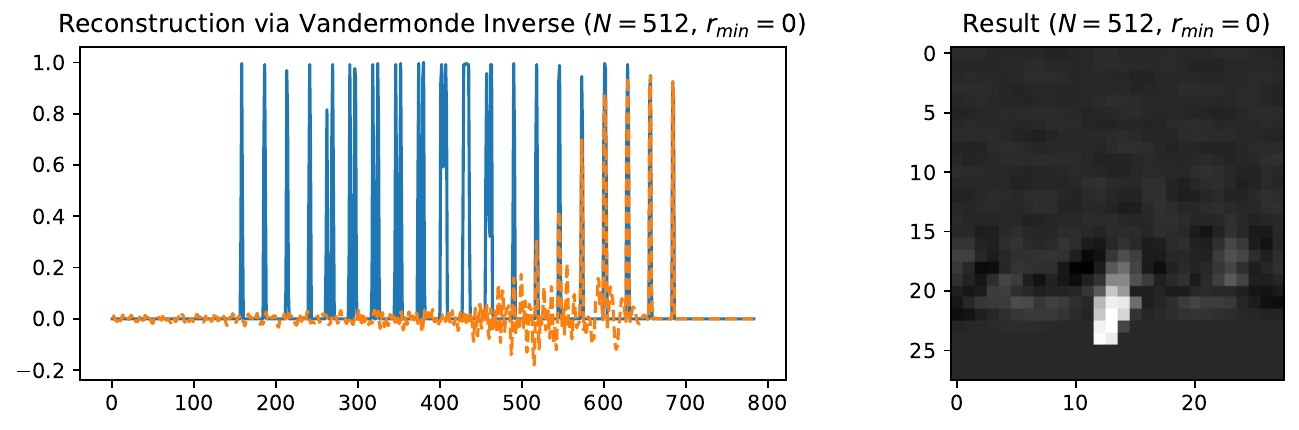}\hspace{10mm}
    \includegraphics[height=0.14\textwidth]{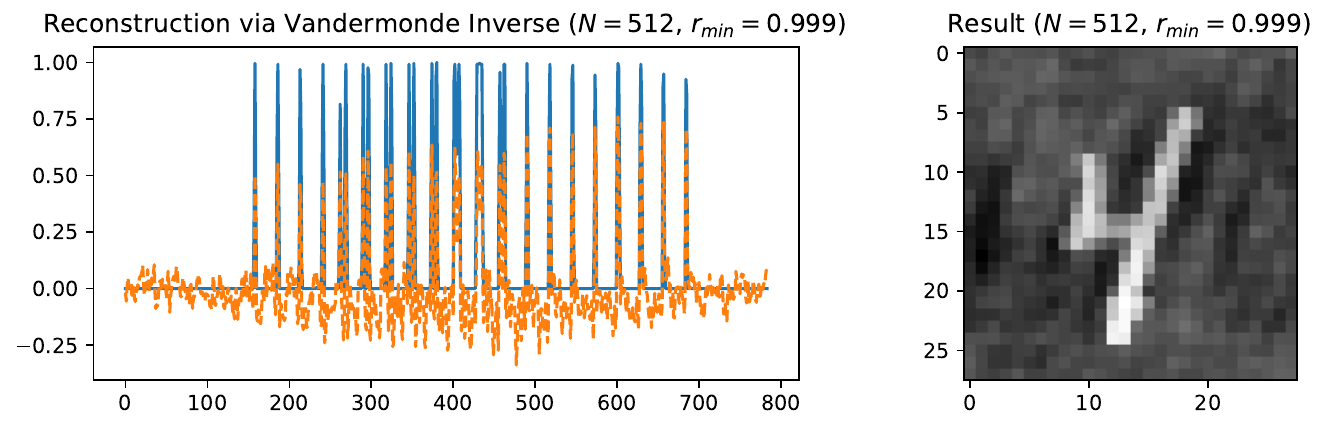}\\
    \includegraphics[height=0.14\textwidth]{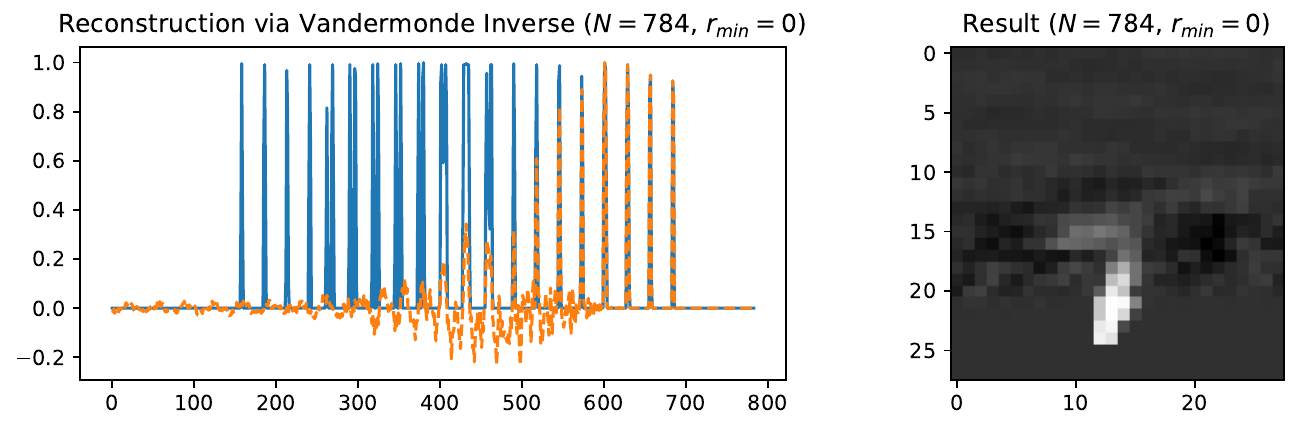}\hspace{10mm}
    \includegraphics[height=0.14\textwidth]{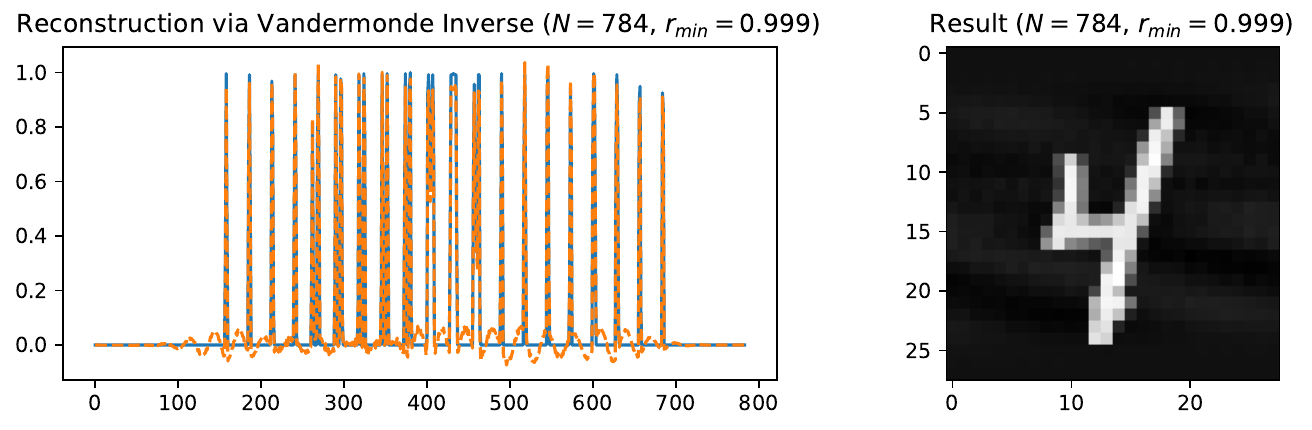}\\
    \includegraphics[height=0.14\textwidth]{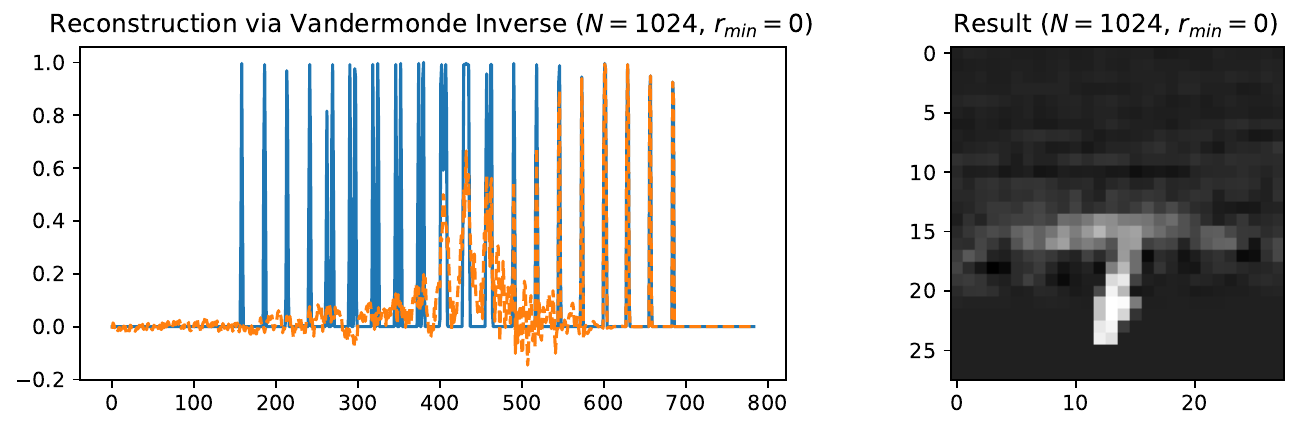}\hspace{10mm}
    \includegraphics[height=0.14\textwidth]{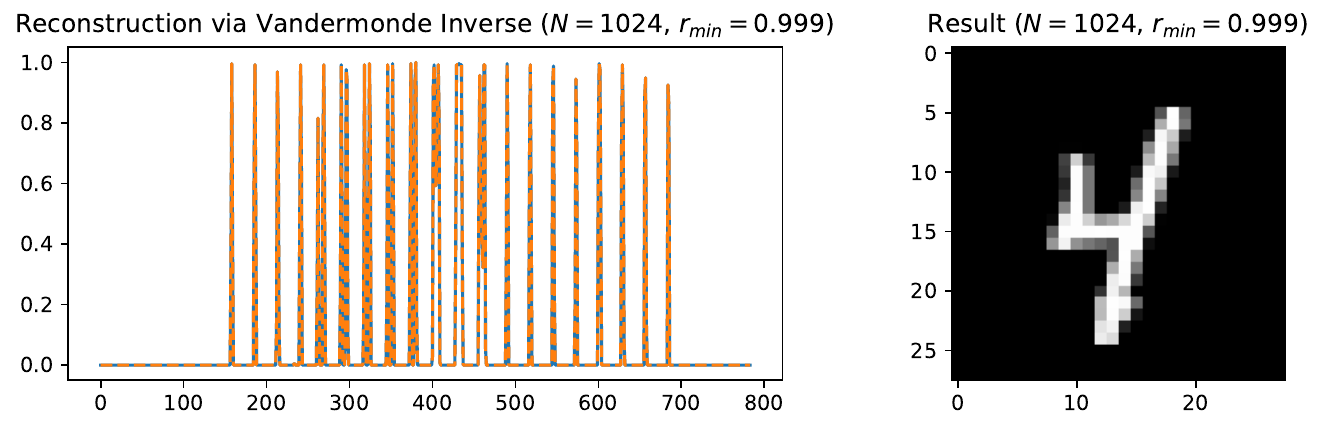}
    \caption{Reconstrucion of MNIST digits~(seen as a sequence) from the last linear RNN state. \textbf{Reconstruction map is not learned} but is instead based on the Vandermonde Pseudoinverse~(i.e. worst-case setting, as discussed in the main text). Results are far more accurate if the reconstruction map is learned~(see Fig.~\ref{fig:MNIST_PF_rec_MLP}).}
    \label{fig:MNIST_rec_random_van}
\end{figure}

In Figure~\ref{fig:MNIST_rec_random_van} we provide evidence on the MNIST dataset~(flattened image $=$ 784 tokens): we observe that, if eigenvalues are sampled with $r_{\min}=0$, by multiplying $x_L$ by $V_L^+$ we are only able to recover recent history. Instead, moving closer to the unit disk allows almost perfect reconstruction at $N=784$, and a satisfactory result already at $N=512$.

\begin{figure}[ht]
    \centering
    \includegraphics[height=0.25\textwidth]{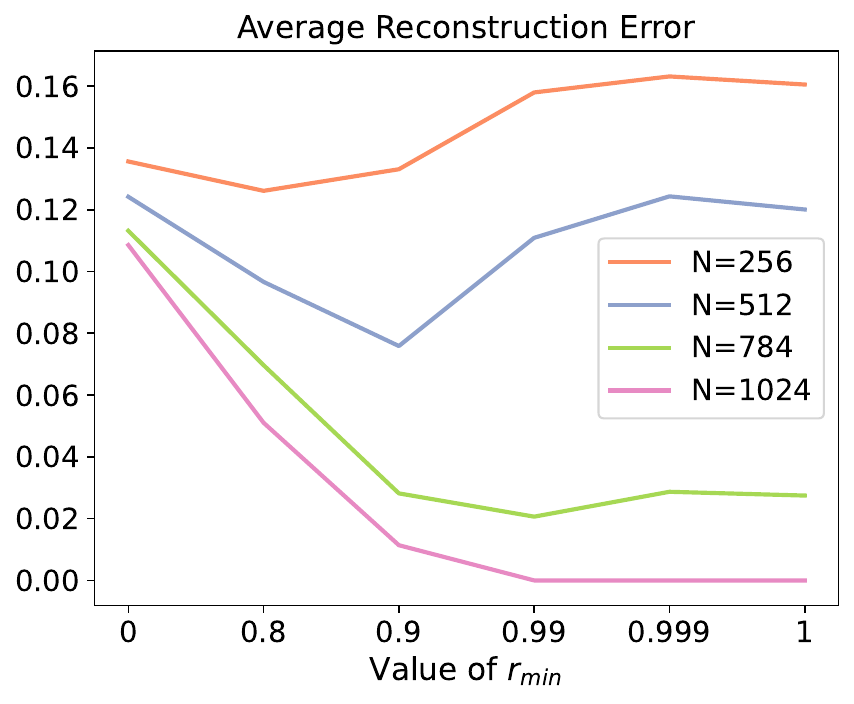}
    \includegraphics[height=0.25\textwidth]{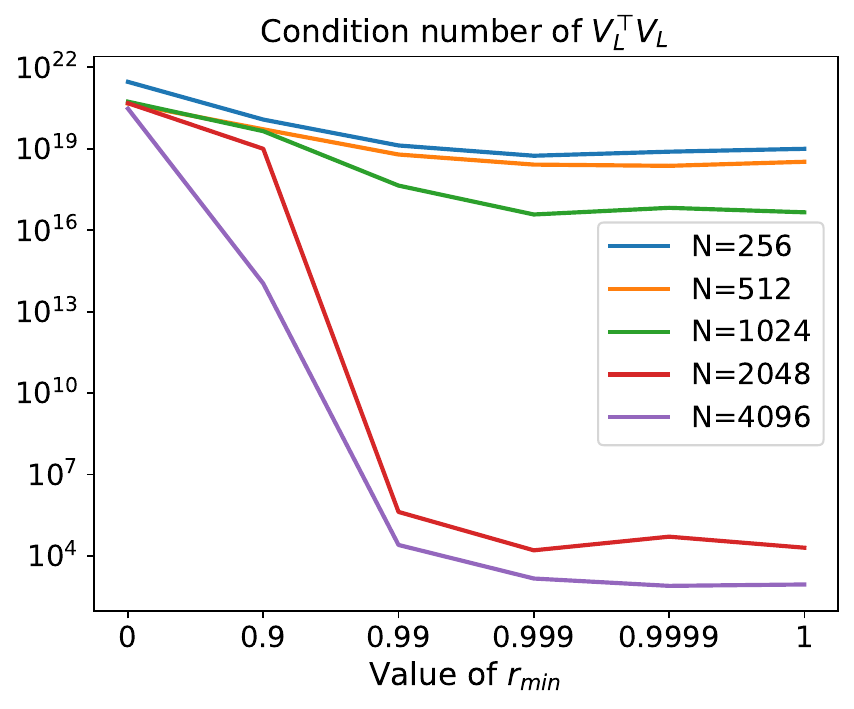}    \\
    \includegraphics[height=0.25\textwidth]{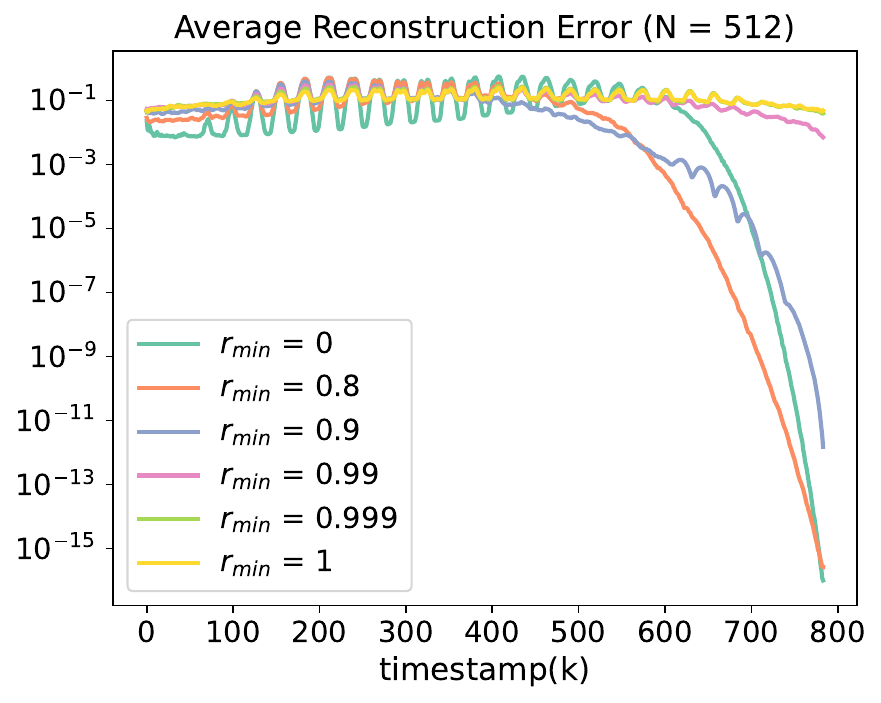}
    \includegraphics[height=0.25\textwidth]{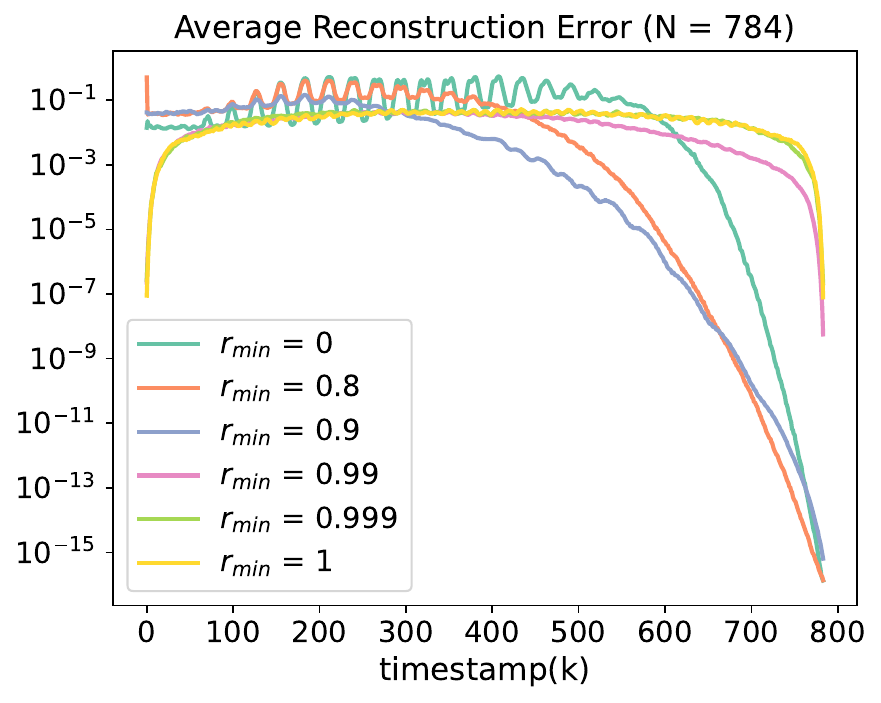}
    \includegraphics[height=0.25\textwidth]{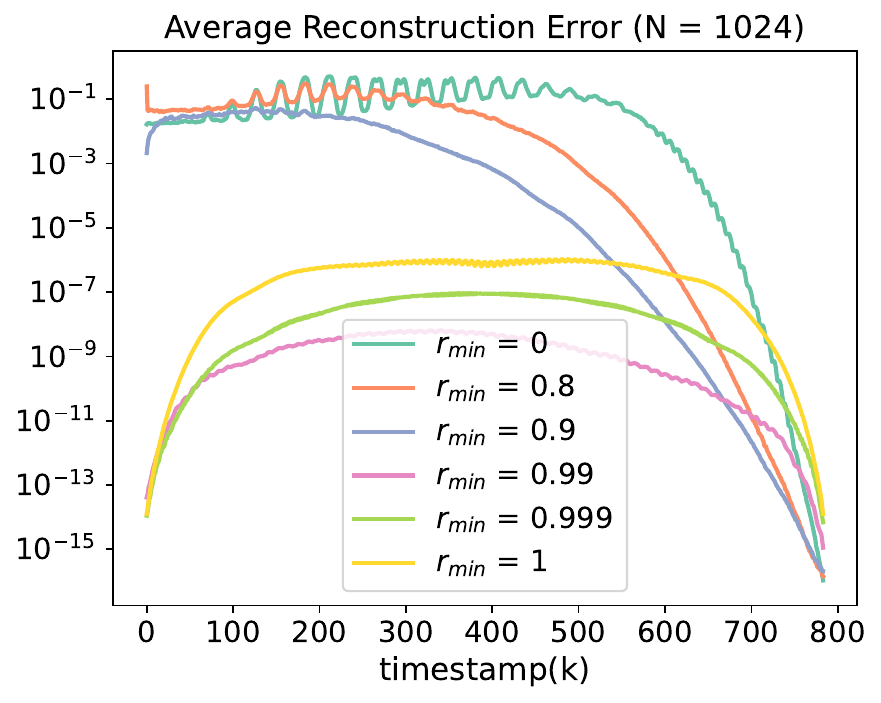}
    \caption{Error over $10k$ MNIST images and 10 re-sampling of the RNN. Comment in text. Reconstruction is based on the Vandermonde Pseudoinverse.}
    \label{fig:MNIST_van_error_avg}
\end{figure}

 In Figure~\ref{fig:MNIST_van_error_avg} we clearly show that the average reconstruction error~(average over $10k$ images samples and 10 random re-samplings of the linear RNN) is decreasing both as a function of the hidden state size~(see discussion in Sec.~\ref{sec:vandermonde}) and of $r_{\min}$~($r_{\max}=1$). The same pattern is observed for the condition number of $V_L^\top V_L$. On the same figure, we show how the error is distributed over timestamps: it is clear that, for $r_{\min}\ll1$, the reconstruction only covers the last few hundreds of tokens -- a property which is liked to the bad condition number observed in this setting.

\begin{figure}[ht]
    \centering
    \includegraphics[height=0.15\textwidth]{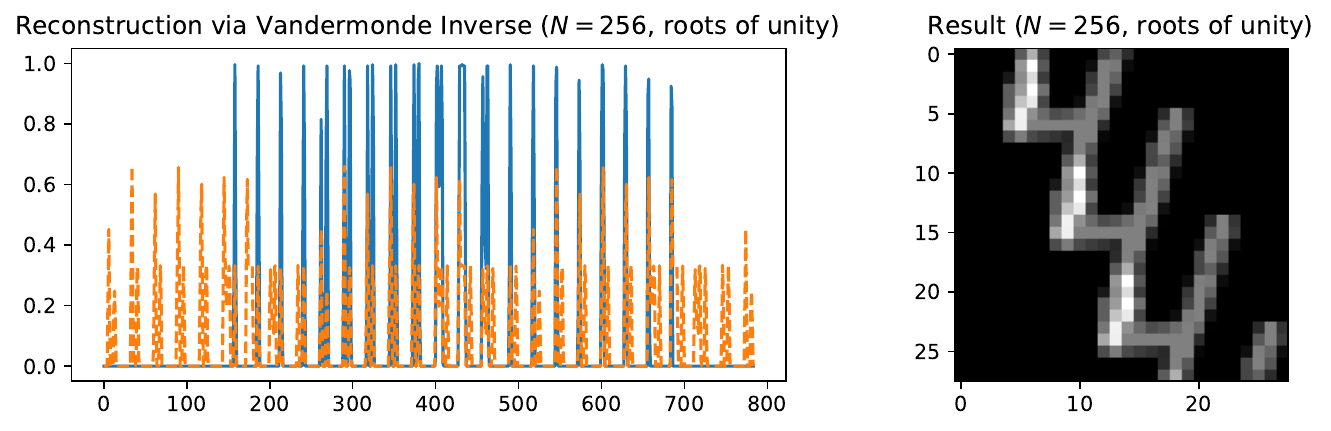}\hspace{10mm}
    \includegraphics[height=0.15\textwidth]{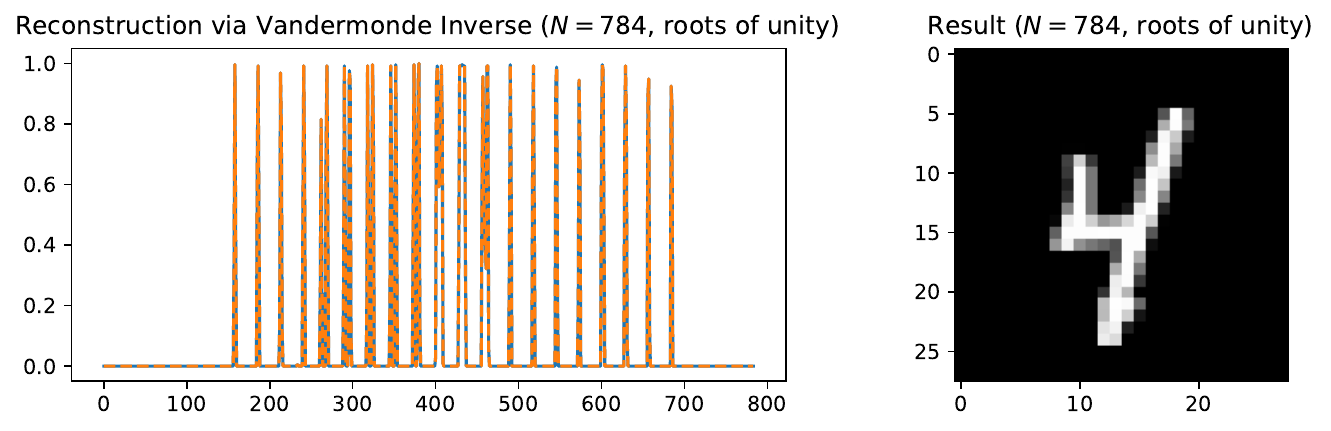}
    \caption{Same setting as Figure~\ref{fig:MNIST_rec_random_van}. Eigenvalues are the $N$-th complex roots of $1$. Comment in text. \textbf{Reconstruction map is not learned} but is instead based on the Vandermonde Pseudoinverse~(i.e. worst-case setting, as discussed in the main text).}
    \label{fig:MNIST_rec_unit_van}
\end{figure}

 Last, in Figure~\ref{fig:MNIST_rec_unit_van} we show what happens when picking the $N$ diagonal entries of $\Lambda$ to be the $N$-th complex roots of $1$: as shown in~\cite{cordova1990vandermonde}, in this setting the Vandermonde condition number is $1$. We observe that we can indeed reconstruct perfectly the output for $N=784$. However, for smaller values of $N$, the reconstruction presents undesired artifacts.

\subsection{Reconstruction under sparsity~(support to \S\ref{sec:sparse})}
\label{app:rec_sparse}

\begin{figure}
    \centering
    \includegraphics[height=0.26\textwidth]{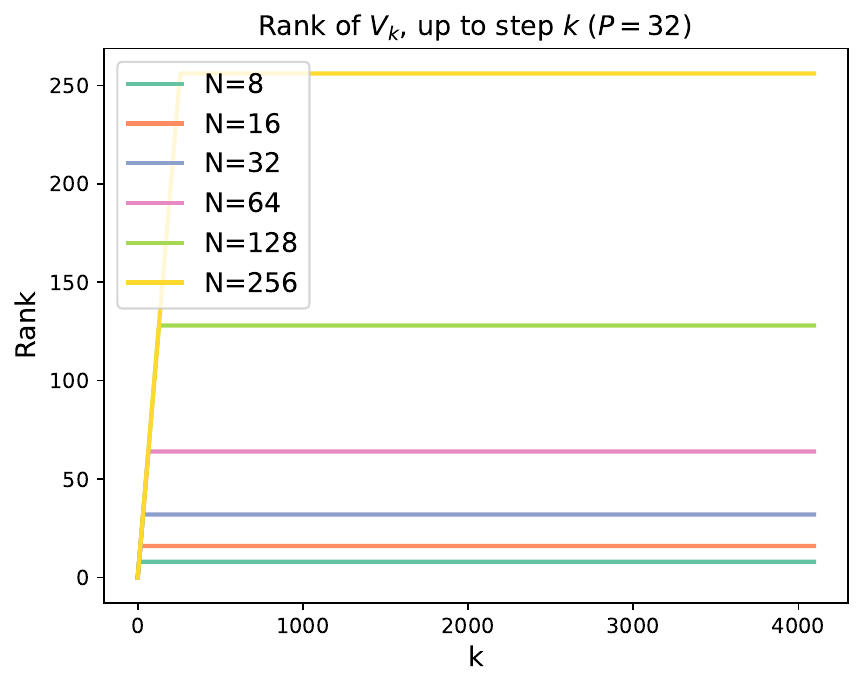}
    \includegraphics[height=0.26\textwidth]{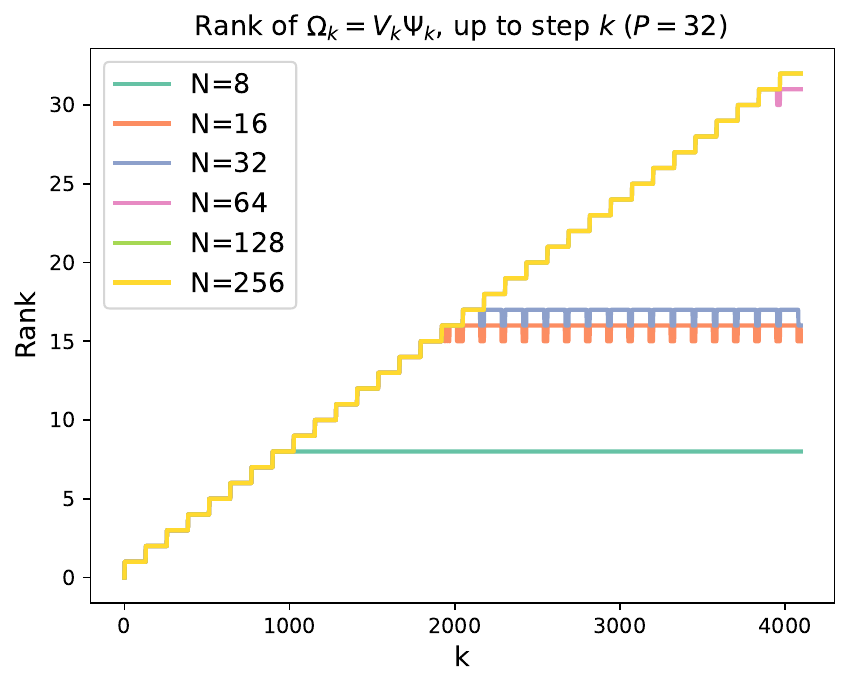}
    \includegraphics[height=0.26\textwidth]{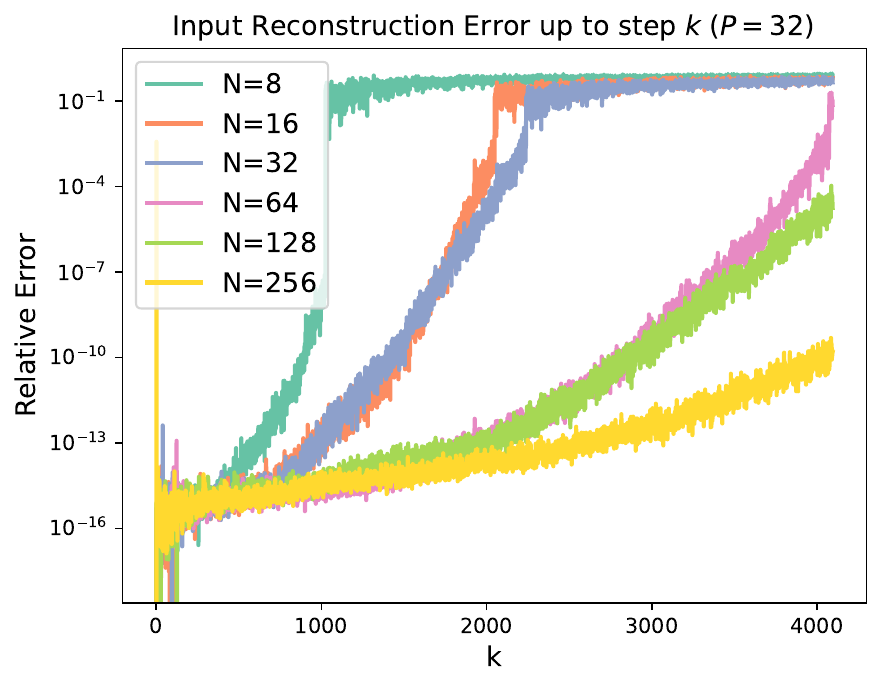}
    \caption{Reconstruction of a random sparse input. $\Lambda$ initialized uniformly at random on $\mathbb{T}(0.95,1)$. Comment in the text.}
    \label{fig:prop1}
\end{figure}

\begin{figure}
    \centering
    \includegraphics[height=0.26\textwidth]{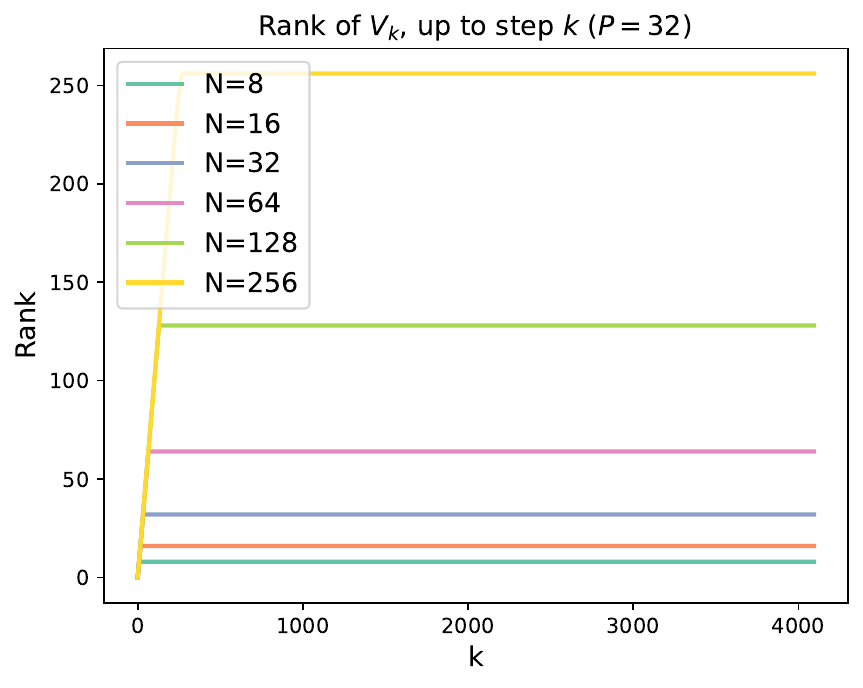}
    \includegraphics[height=0.26\textwidth]{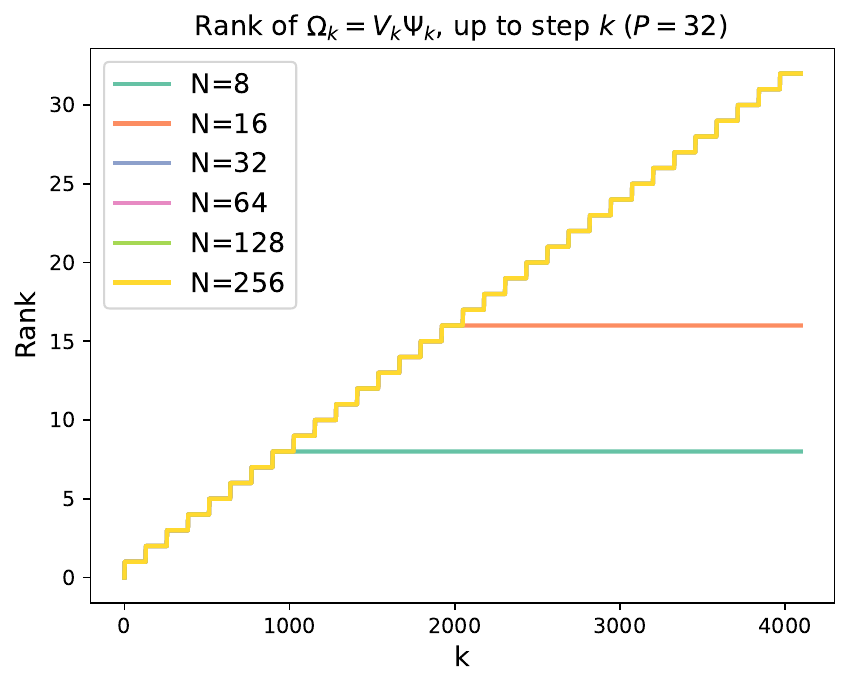}
    \includegraphics[height=0.26\textwidth]{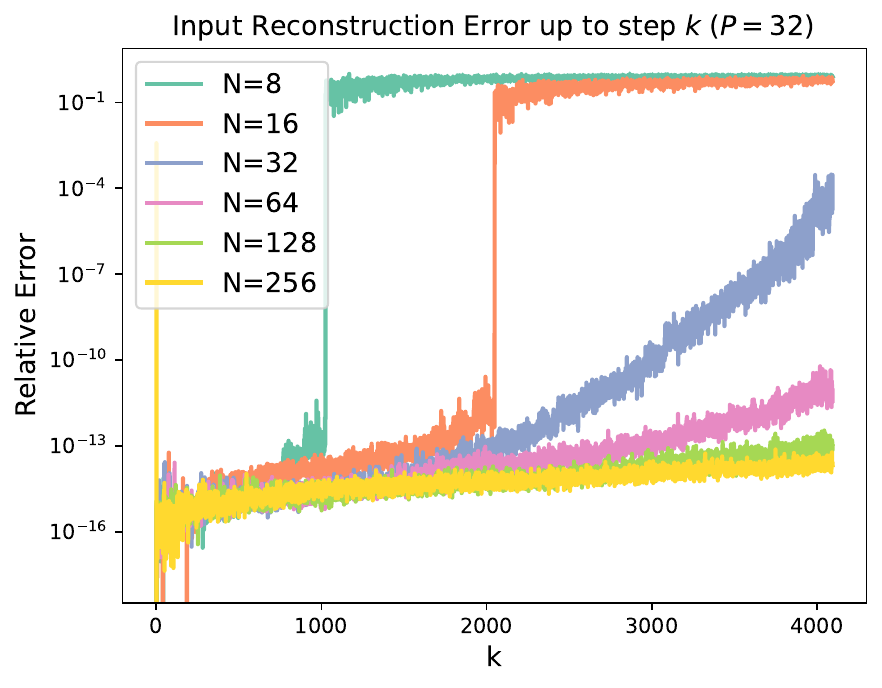}
    \caption{Reconstruction of a random sparse input. $\Lambda$ initialized uniformly at random on $\mathbb{T}(0.99,1)$. Comment in the text.}
    \label{fig:prop2}
\end{figure}

In Figures~\ref{fig:prop1} and \ref{fig:prop2} we test the discussion in Sec.~\ref{sec:sparse} in a controlled setting. We consider one-dimensional stream of $L=4096$ random inputs sparse in a basis of $P=32$ Haar wavelets~\citep{haar1911theorie}. The linear diagonal RNN has $\Lambda\in\C^{N\times N}$ with eigenvalues sampled uniformly at random from $\mathbb{T}(0.95,1)$~(Fig.~\ref{fig:prop1}) or $\mathbb{T}(0.99,1)$~(Fig.~\ref{fig:prop2}), where
$$\mathbb{T}[r_{\min}, r_{\max}] := \{\lambda\in\C \ | \ r_{\min}\le|\lambda|\le r_{\max}\}.$$
We use matrix $B = (1,1,\dots, 1)^\top$. Plotted is the rank of $V_k,\Psi_k$, $\Omega_k = V_k\Psi_k$ as $k$ increases~(see notation in Sec.~\ref{sec:sparse}). We show how the reconstruction error behaves when reconstructing $u_{1:k} = \Psi_k\alpha^{u}_k$ with $\alpha^{u}_k = \Omega_k^+ x_k$. In the figures, we plot the error for reconstruction of the tokens $(u_i)_{i=1}^k$ from $x_k$, for all $k\le L$. As $N$ gets larger the reconstruction error gets uniformly negligible. In particular, if we initialize in $\mathbb{T}(0.95,1)$ then the minimum $N$ we need for perfect reconstruction of around $N=256$. If instead we initialize closer to the unit circle, then $N=64$ is enough for perfect reconstruction. This finding is similar to the one presented in Sec.~\ref{app:rec_vandermonde}. 

\begin{figure}
    \centering
    \includegraphics[height=0.25\textwidth]{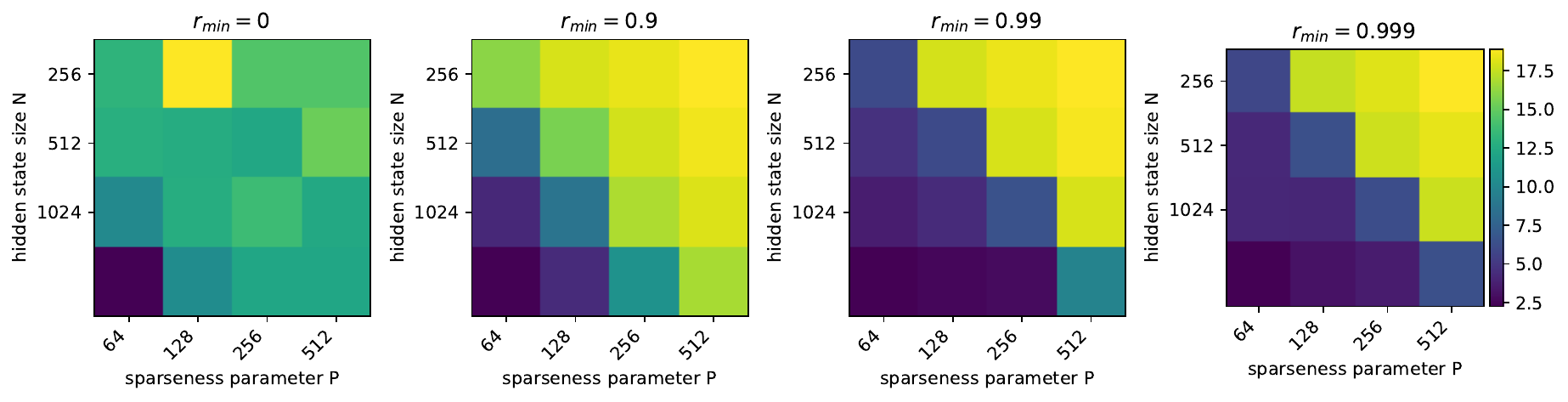}
    \caption{Logarithm of the condition number for $\Omega^T\Omega$ decreases as $N$~(hidden state dimension) increases and as $P$~(number of basis functions) decreases. As always, $r_{\min}$ closer to 1 leads to better conditioning and therefore more stable reconstruction.}
    \label{fig:better_conditioning}
\end{figure}

On a more fundamental level, we study the condition number of the matrix $\Omega^T\Omega$, where $\Omega = V_L\Psi_L$. This condition number quantifies the numerical stability of the pseudoinverse $\Omega^+$, used in reconstruction. In Figure~\ref{fig:better_conditioning}, we show the logarithm of the condition number for a sequence of length $512$ sparse in a basis of $P$ eigenfunctions. As $r_{\min}$ gets close to $1$ the condition number decreases as we saw in \S\ref{app:rec_vandermonde}. Crucially however, the condition number also decreases as $P$ decreases.

\begin{figure}[h]
    \centering
    \includegraphics[width=0.5\linewidth]{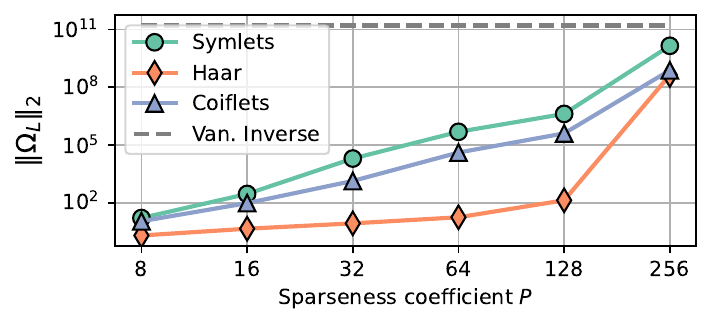}
    \vspace{-3mm}
    \caption{Conditioning of the reconstruction map for inputs sparse in 3 different wavelet basis~($N=256$). Average over $1k$ random init., $r_{\min}=0.999, r_{\max}=1$. Comment in the main text.}
    \vspace{-3mm}
    \label{fig:effect_P}
\end{figure}

\subsection{Approximation of sequence-to-sequence maps~(ODE Systems)}
We consider approximating sequence-to-sequence maps $(v_i)_{i=1}^L\overset{T}{\mapsto} (y_i)_{i=1}^L$ defined by Runge-Kutta discretization of the flow of a controlled differential equation $\dot z_t = f(z_t, v_t), y_t = h(z_t)$, where $(v_t)_{t}$ is the input, $f$ is a non-linear multidimensional function, $h$ projects the multidimensional state $z_t$ into a one-dimensional output. An example is the \textbf{Protein Transduction~(PT)} system~\citep{vyshemirsky2008bayesian}:
 \begin{align*}
     &\dot z_1(t) = -k_1 z_1(t) - k_2 z_1(t) z_3(t) + k_3 z_4(t) + v(t)\\
     &\dot z_2(t) = k_1 z_1(t)\\
     &\dot z_3(t) = - k_2 z_1(t) z_3(t) + k_3 z_4(t) + V \frac{z_5(t)}{K_m + z_5(t)}\\
     &\dot z_4(t) = k_2 z_1(t) z_3(t) - (k_3+k_4)z_4(t)\\
     &\dot z_5(t) = k_4 z_4(t) - V\frac{z_5(t)}{K_m + z_5(t)}
 \end{align*}
\vspace{-2mm}

We identify $y_k = z_1(\Delta k)$, where $\Delta = 0.01$, and $v_k = v(\Delta k)$. We sample $(v_i)_{i=1}^L$~($L=2048$ in PT) from a linear combination of 16 base wavelets of low frequency (\textit{bias: input is random but has slow variations}), and set the ground truth $(y_i)_{i=1}^L$ to be the result of Runge-Kutta integration with stepsize $\Delta$. As hyperparameters, we use $k_1 = 0.07, k_2 = 0.6, k_3 =
0.05, k_4 = 0.3, V = 0.017, K_m = 0.3, z_1(0)=1, z_2(0) = 0, z_3(0) = 1, z_4(0) =  z_5(0) = 0$ as prescribed by~\citet{vyshemirsky2008bayesian}. Results using approximation of one linear RNN followed by a 1HL-MLP~(shared across timestamps) are shown in Fig.~\ref{fig:ptpz} and discussed in the main text. To train, we use $10k$ random (low frequency) trajectories and test on $1k$ trajectories with same distribution. Experiments run on a single A5000 using an LRU in JAX~(\url{https://github.com/NicolasZucchet/minimal-LRU/tree/main/lru}).

\begin{figure}[ht]
    \centering
    \includegraphics[width=0.99\linewidth]{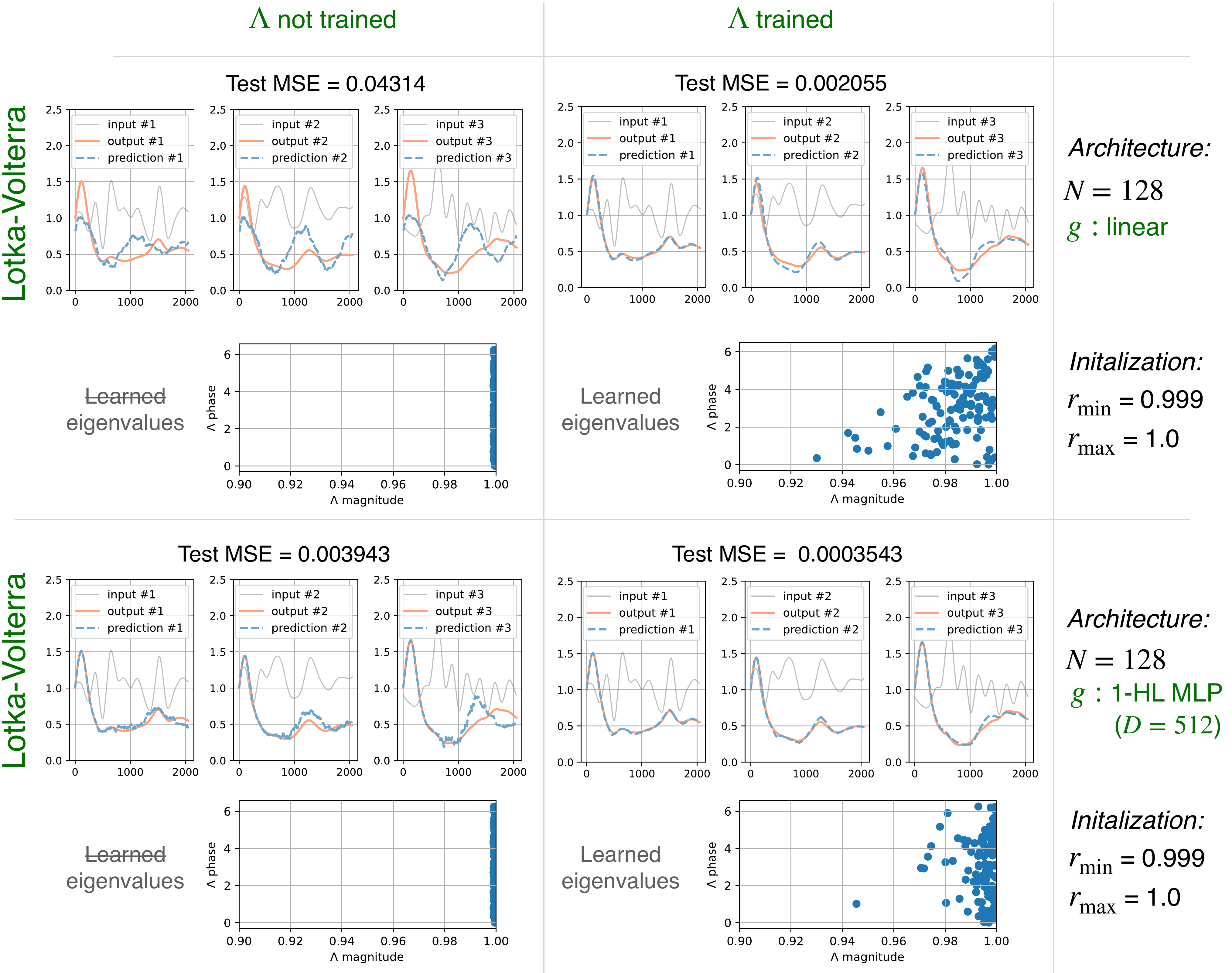}
    \vspace{-2mm}
    \caption{Performance of one layer linear RNN followed by a linear layer or a 1HL-MLP~(shared accross timestamps), when training or clamping the RNN at initialization. The encoder and the readout from the RNN hidden state are always trained. \textbf{Results:} The model greatly benefits from learning the recurrent eigenvalues, but works already quite well with linear projections on the hidden state -- indicating the LV system can be approximated (with some noticeable yet small error) by a linear functional, i.e. a linear system~(see Thm~\ref{thm:linear_rnn}). When used, the 1HL-MLP has $D = 512$ hidden neurons, to double the input size of $2N$~(real + imaginary part). Even though some elements of our theory suggest $r_{\min}\simeq 1$ for reconstruction guarantees, here we clearly see that successful learning pushes eigenvalues slightly inside the unit disk. As~\citep{gu2021efficiently, smith2022simplified, orvieto2023resurrecting} we found it benefitial to initialize eigenvalues close to the unit circle.}
    \label{fig:res_lv}
\end{figure}

In this appendix, we additionally discuss performance in approximating the solution of two other controlled ODEs. Settings are same as used for PT unless stated otherwise. The first ODE~(results in Fig.~\ref{fig:res_lv}) is a \textbf{Lotka-Volterra (LV)} system~\citep{lotka1925elements, volterra1928variations}:
\begin{align*}
\dot z_1(t) &= z_1(t)(a-b\cdot z_2(t)) \\
\dot z_2(t) &= -z_2(t)(c-d \cdot z_1(t))+v(t)
\end{align*}
where $a = 1.0, b= 0.6, c=1.0, d=0.7$, and we initialize $z_1(0) = 1.0 , z_2(0) = 0.5$ as used in~\citet{dondelinger2013ode}. For integration, we use a stepsize of $\Delta =0.01$ and $y_k = z_1(\Delta k)$. Here, sequence length is again $L=2048$.

\begin{figure}[ht]
    \centering
    \includegraphics[width=0.3\linewidth]{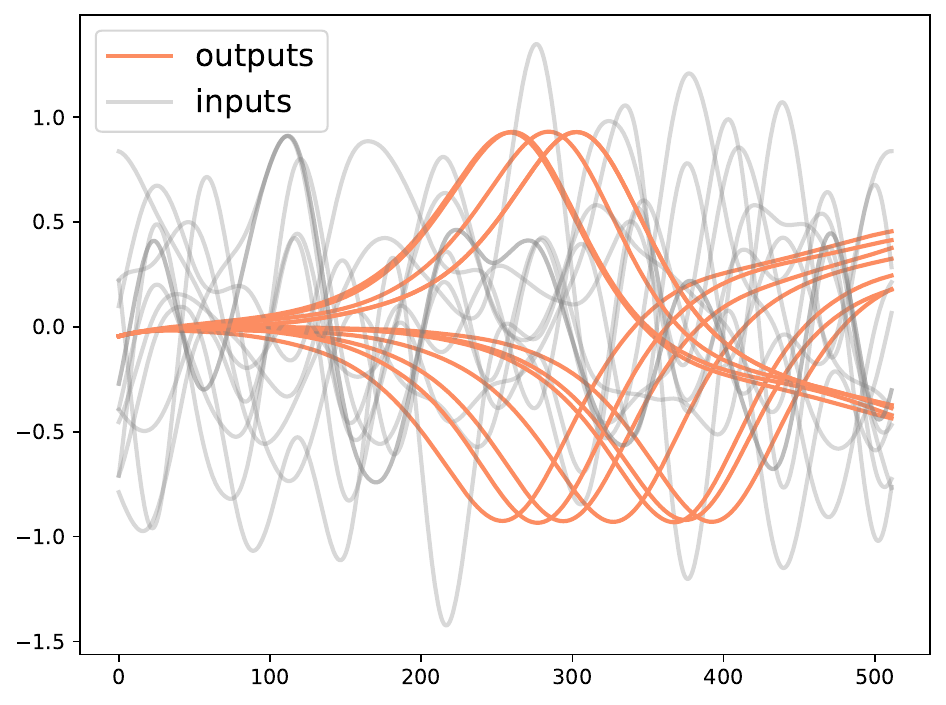}\hspace{15mm}
    \includegraphics[width=0.3\linewidth]{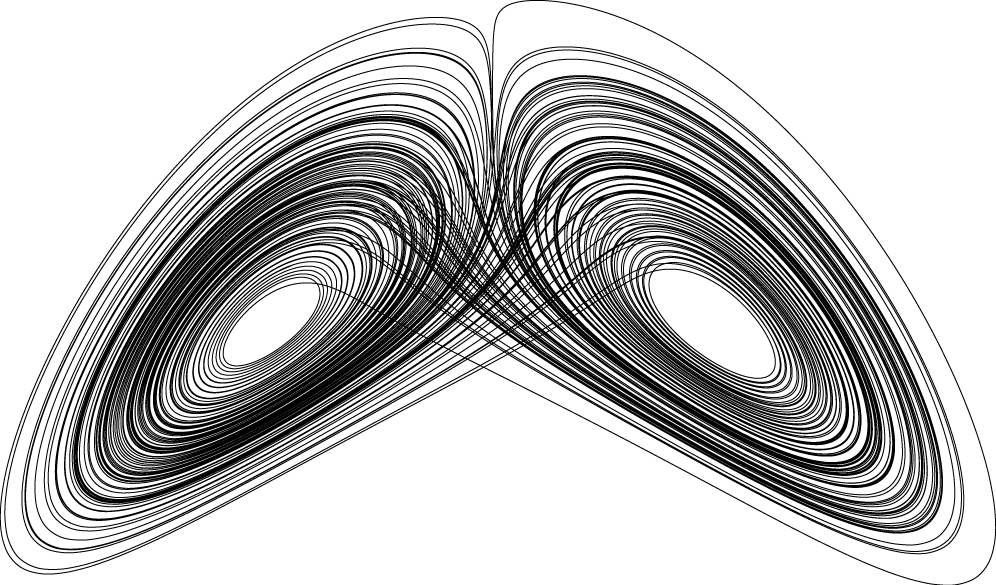}
        \vspace{-2mm}
    \caption{Behavior of the input-output LV map for random low-frequency inputs controlling the second state equation~(output is the discretized first component). The output has binary nature~(up or down) with shifted phase: tiny variations in the input cause drastic effects on the output~(butterfly effect). Plotted is also an illustration of the 3-dimensional dynamics, source: \url{https://www.iaacblog.com/programs/algorithmic-emergence-chaotic-attractor-equations/}.}
    \label{fig:lorentz_ill}
\end{figure}

\begin{figure}[ht]
    \centering
    \includegraphics[width=0.4\linewidth]{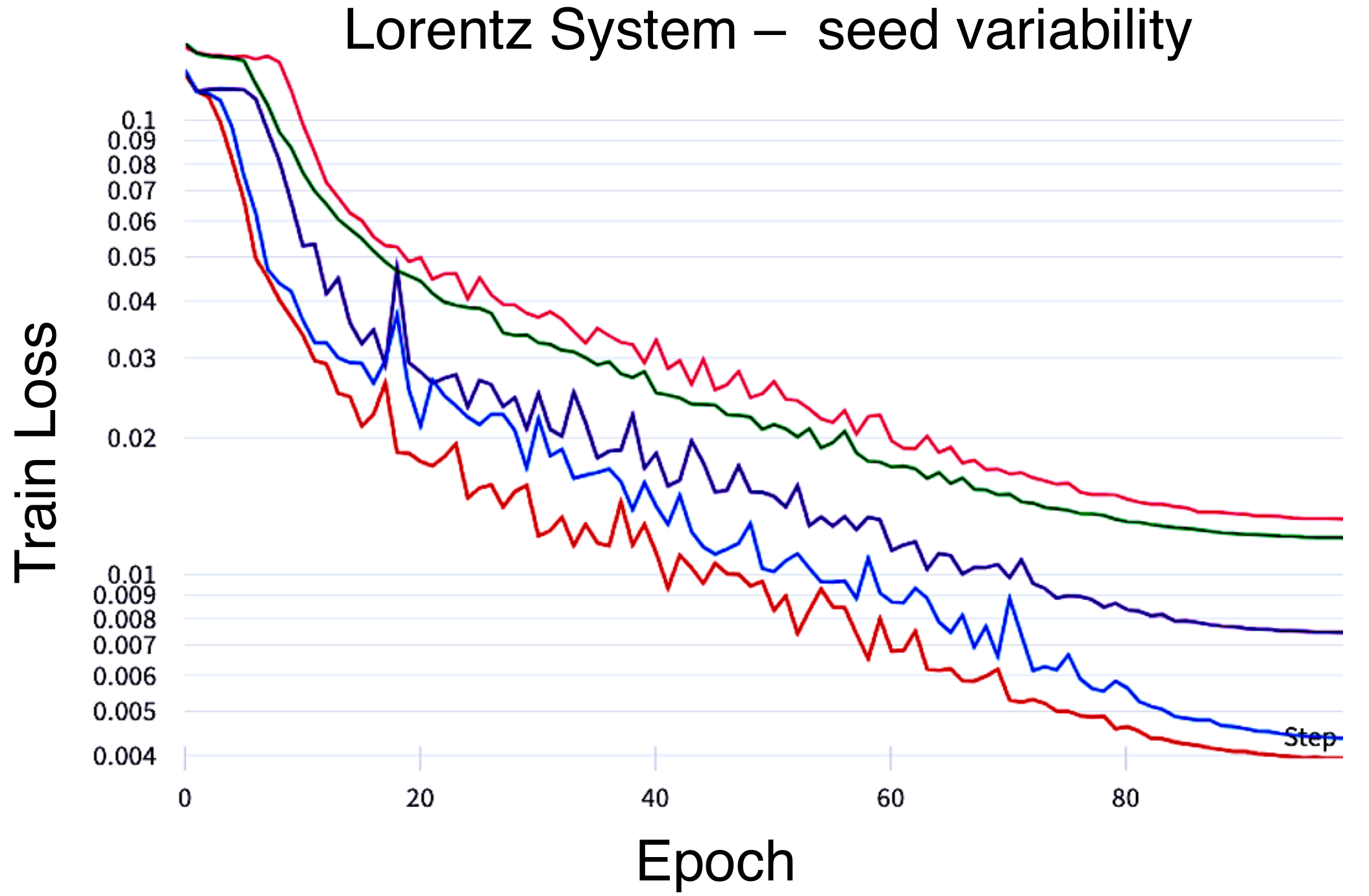}
        \vspace{-2mm}
    \caption{Variability in training loss dynamics when dealing with randomly initialized linear RNN + 1HL-MLP. Plotted are the dynamics for seeds $1-6$. All other hyperparameters are shared across runs.}
    \label{fig:seed_variability}
\end{figure}

\begin{figure}[ht]
    \centering
    \includegraphics[width=0.99\linewidth]{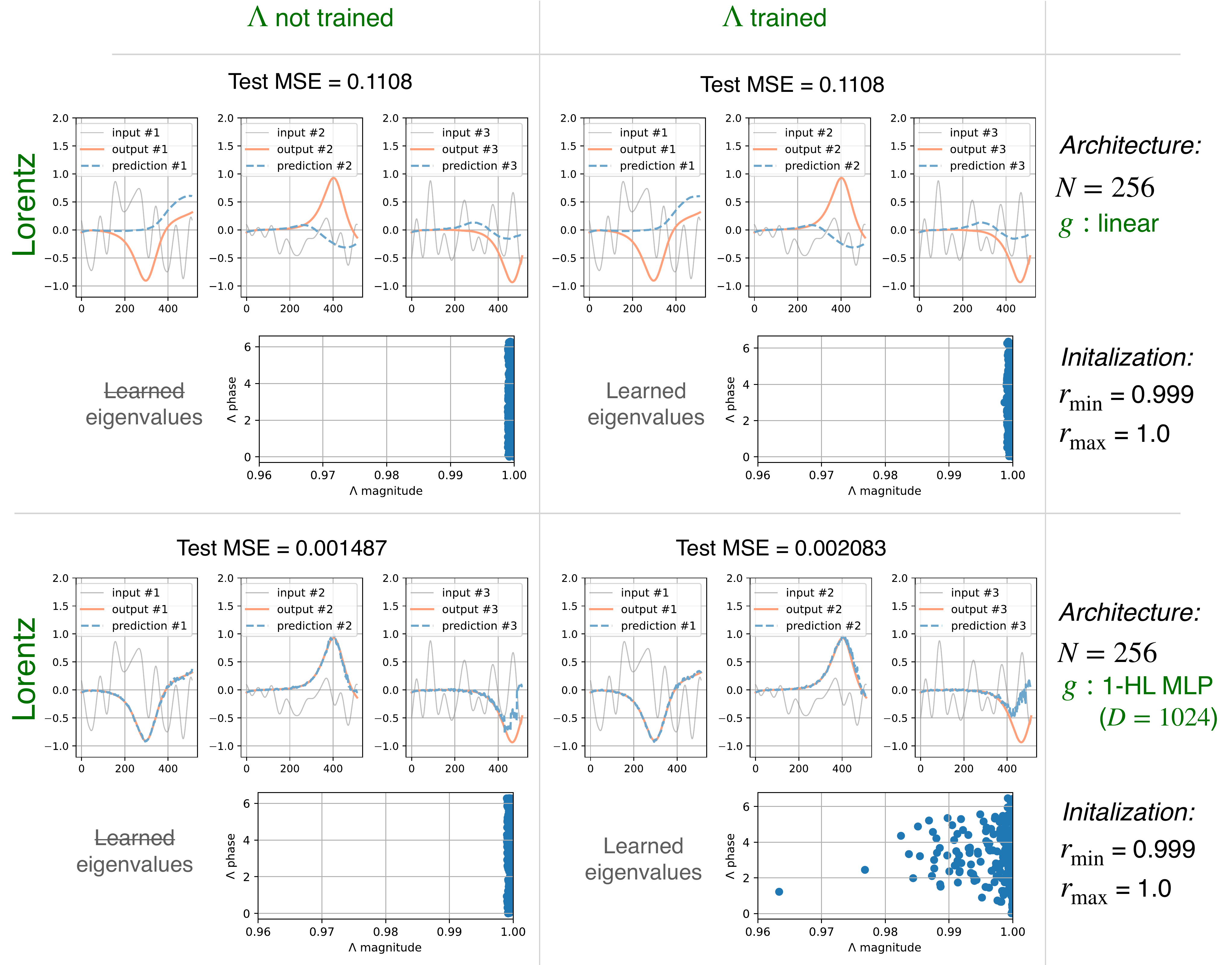}
        \vspace{-2mm}
    \caption{\textbf{Results:} compared to Fig.~\ref{fig:res_lv}, here the task is considerably more challenging: no linear functional is able to learn useful information about the system --- the 1HL-MLP is necessary for decent approximation as the system is highly non-linear. Learning input-output maps with $L=512$ is already challenging: a larger~(compared to LV) hidden state $N=256$ and an MLP width $D=1024$ are therefore chosen for this task. The learned parameters lead to a solution which is not perfect: at the edge of the interval the learned model struggles. This can be improved with bigger models. However, the learned architecture is able to capture most of the system's chaotic behavior. Surprisingly, compared to LV, we found that here learning just the 1HL-MLP already gives great results: we think this is due to the fact that the sequence length is shorter and the hidden dimension is bigger: the model does not need to adjust eigenvalues as enough variability is provided in the hidden state.}
    \label{fig:res_lz}
\end{figure}

Finally, we consider an extremely \textit{challenging scenario}: the \textbf{Lorentz system~(LZ)}~\citep{lorenz1963deterministic}, which notoriously has chaotic solutions~(named strange attractor, linked to the butterfly effect):
\begin{align*}
\dot z_1(t) &= \sigma \cdot (z_2(t)-z_1(t)) \\
\dot z_2(t) &= (r-z_3(t))z_1(t)-z_2(t)+ v(t) \\
\dot z_3(t) &= z_2(t)z_1(t)-b\cdot z_3(t)
\end{align*}
With our parameter choices $\sigma = 10, r= 26, b =8/3$ and initialization $z_1(0) = -0.89229143 , z_2(0) = 1.08417925, z_3(0) = 2.34322702$~(parameter source: Wikipedia, visited September 2023), the system has a chaotic behavior as shown in Figure~\ref{fig:lorentz_ill}. We choose an integration timestep $\Delta = 0.002$ and, due to the non-linear chaotic and possibly unstable nature of the controlled attractor, consider $L=512$ in this setting.
\vspace{-3mm}
\paragraph{Note on training.} Training one layer of linear RNN + MLP on these ODEs exhibits huge variation across seeds~(see Fig.~\ref{fig:seed_variability}). Since in this paper we want to show that \textit{there exist} a Linear RNN+MLP configuration able to model non-linear sequence to sequence maps, we consider the following setup: we run each experiment and hyperparameter sweep on seeds $1-6$, and only report the best performance on the test data (train loss always lower). Based on our experience with the LRU, we conclude that instability is due to the small dimension of our model: performance on standard tasks is more stable as depth increases~\citep{orvieto2023resurrecting}. To test our theory, we limit ourselves to a linear RNN with either $N=128$ or $N=256$, followed by an MLP with one hidden layer. We grid-search hyperparameters for each model configuration and report test error for the best-performing models. Usual best-performing stepsizes are $0.003$ and $0.01$. In Lotka-Volterra experiments~(Fig.~\ref{fig:res_lv}) we train for $200$ epochs, while we train on the Lorentz System~(Fig.~\ref{fig:res_lz}) for $1000$ epochs. No weight decay, dropout or normalizations are applied. \textbf{Results are discussed in the relative figure captions and nicely validate our claims}.  \textit{Code for reproducing the experiments will be provided upon acceptance of this paper}.

\end{document}